\documentclass{article}

\usepackage{microtype}
\usepackage{graphicx}
\usepackage{booktabs} 

\usepackage{graphicx,amsmath,amsfonts,amssymb,algorithmic,algorithm,cite,caption}
\usepackage{url,epsfig,epsf,xcolor,MnSymbol,mathbbol,fmtcount,multirow}
\usepackage{subfig}
\usepackage{mathtools}
\usepackage{latexsym}
\usepackage{tabularx}
\usepackage{caption}
\usepackage[bottom,hang,flushmargin]{footmisc}
\usepackage{comment}
\usepackage[bottom]{footmisc}
\usepackage{caption}
\usepackage{enumitem}
\usepackage{listings}
\usepackage{comment}

\usepackage{hyperref}

\usepackage[accepted]{icml2019}

\newcommand{\EE}{\mathbb{E}}
\newcommand{\PP}{\mathbb{P}}

\newcommand{\bWW}{\mathbf{W}}

\newcommand{\bb}{\mathbf{b}}
\newcommand{\bx}{\mathbf{x}}
\newcommand{\by}{\mathbf{y}}

\newcommand{\bp}{\mathbf{p}}

\newcommand{\bv}{\mathbf{v}}

\newcommand{\bg}{\mathbf{g}}
\newcommand{\bu}{\mathbf{u}}

\newcommand{\bA}{\mathbf{A}}
\newcommand{\bB}{\mathbf{B}}
\newcommand{\bC}{\mathbf{C}}
\newcommand{\bU}{\mathbf{U}}
\newcommand{\bV}{\mathbf{V}}

\newcommand{\bL}{\mathbf{L}}

\newcommand{\bI}{\mathbf{I}}

\newcommand{\bz}{\mathbf{z}}

\newcommand{\bmu}{\mathbf{\mu}}

\newcommand{\bH}{\mathbf{H}}

\newcommand{\tbx}{\tilde{\mathbf{x}}}

\newcommand{\grad}{\nabla}

\newcommand{\cX}{\mathcal{X}}
\newcommand{\cY}{\mathcal{Y}}

\newcommand{\cO}{\mathcal{O}}

\newcommand{\cC}{\mathcal{C}}

\newcommand{\cS}{\mathcal{S}}

\newcommand{\tI}{\tilde{I}}
\newcommand{\eps}{\epsilon}

\newcommand{\hby}{\hat{\by}}

\newcommand{\bSigma}{\mathbf{\Sigma}}
\newcommand{\bsigma}{\mathbf{\sigma}}

\def\<{\langle}
\def\>{\rangle}

\newcommand{\name}{CASO}
\newcommand{\firstorder}{CAFO}

\DeclareMathOperator*{\argmin}{arg\,min}

\newtheorem{theorem}{\textbf{Theorem}}

\newtheorem{corollary}{\textbf{Corollary}}

\newtheorem{definition}{\textbf{Definition}}

\newtheorem{proposition}{\textbf{Proposition}}

\newcommand{\tell}{\tilde{\ell}}

\def \endprf{\hfill {\vrule height6pt width6pt depth0pt}\medskip}

\icmltitlerunning{Understanding High-Order Loss Approximations and Features in Deep Learning Interpretation}

\begin{document}

\twocolumn[
\icmltitle{Understanding Impacts of High-Order Loss Approximations and Features in Deep Learning Interpretation}



\icmlsetsymbol{equal}{*}

\begin{icmlauthorlist}
\icmlauthor{Sahil Singla}{to}
\icmlauthor{Eric Wallace}{to}
\icmlauthor{Shi Feng}{to}
\icmlauthor{Soheil Feizi}{to}
\end{icmlauthorlist}

\icmlaffiliation{to}{Computer Science Department, University of Maryland}

\icmlcorrespondingauthor{Sahil Singla}{ssingla@cs.umd.edu}
\icmlcorrespondingauthor{Soheil Feizi}{sfeizi@cs.umd.edu}

\icmlkeywords{Deep Learning Interpretation, Hessian, $L_1$ regularization}

\vskip 0.3in
]



\printAffiliationsAndNotice{}  

\begin{abstract}
Current saliency map interpretations for neural networks generally rely on two key assumptions. First, they use first-order approximations of the loss function, neglecting higher-order terms such as the loss curvature. Second, they evaluate each feature's importance in isolation, ignoring feature interdependencies. This work studies the effect of relaxing these two assumptions. First, we characterize a closed-form formula for the input Hessian matrix of a deep ReLU network. Using this, we show that, for classification problems with many classes, if a prediction has high probability then including the Hessian term has a small impact on the interpretation. We prove this result by demonstrating that these conditions cause the Hessian matrix to be approximately rank one and its leading eigenvector to be almost parallel to the gradient of the loss. We empirically validate this theory by interpreting ImageNet classifiers. Second, we incorporate feature interdependencies by calculating the importance of group-features using a sparsity regularization term. We use an $L_0-L_1$ relaxation technique along with proximal gradient descent to efficiently compute group-feature importance values. Our empirical results show that our method significantly improves deep learning interpretations.
\end{abstract}

\section{Introduction}

The growing use of deep learning in sensitive applications such as medicine, autonomous driving, and finance raises concerns about human trust in machine learning systems. For trained models, a central question is test-time \emph{interpretability}: how can humans understand the reasoning behind model predictions? A common interpretation approach is to identify the importance of each input feature for a model's prediction. A saliency map can then visualize the important features, e.g., the pixels of an image~\cite{simonyan2013deep,sundararajan2017axiomatic} or words in a sentence~\cite{li2016understanding}. 

Several approaches exist to create saliency maps, largely based on model gradients. For example, \citet{simonyan2013deep} compute the gradient of the class score with respect to the input, while \citet{smilkov2017smoothgrad} average the gradient from several noisy versions of the input. Although these gradient-based  methods can produce visually pleasing results, they often weakly approximate the underlying model~\cite{feng2018rawr,nie2018backprop}. Existing saliency interpretations mainly rely on two key assumptions:
\begin{itemize}
	\itemsep0em 

\item {\bfseries Gradient-based loss surrogate:} For computational efficiency, several existing methods, e.g., \citet{simonyan2013deep, smilkov2017smoothgrad, sundararajan2017axiomatic}, assume that the loss function is almost linear at the test sample. Thus, they use variations of the input gradient to compute feature importance. 

\item  {\bfseries Isolated feature importance:} Current methods evaluate the importance of each feature in isolation, assuming all other features are fixed. Features, however, may have complex interdependencies that can be learned by the model. 
\end{itemize}

This work studies the impact of relaxing these two assumptions in deep learning interpretation. To relax the first assumption, we use the second-order approximation of the loss function by keeping the Hessian term in the Taylor expansion of the loss. For a deep ReLU network and the cross-entropy loss function, we compute this Hessian term in {\it closed-form}. Using this closed-form formula for the Hessian, we prove the following for ReLU networks:
\begin{theorem}[informal version]
If the probability of the predicted class is close to one and the number of classes is large, first-order and second-order interpretations are sufficiently close to each other.
\end{theorem}
We present a formal version of this result in Theorem \ref{thm:equivalence} and also validate it empirically. For instance, in ImageNet 2012~\cite{russakovsky2015imagenet}, a dataset of 1,000 classes, we show that incorporating the Hessian term in deep learning interpretation has a small impact for most images.

The key idea of the proof follows from the fact that when the number of classes is large and the confidence in the predicted class is high, the Hessian of the loss function is approximately of rank one. In essence, the largest eigenvalue squared is significantly larger than the sum of squared remaining eigenvalues. Moreover, the corresponding eigenvector is approximately parallel to the gradient vector (Theorem \ref{thm:hessianparallel}). This causes first-order and second-order interpretations to perform similarly. We also show in Appendix~\ref{subsec:nonrelu_networks} that this result holds empirically for a neural network model that is not piecewise linear. Our theoretical results can also be extended to related problems such as adversarial examples, where most methods are based on the first-order loss approximations~\cite{2014arXiv1412.6572G, 2015arXiv151104599M, madry2018towards}.
	
Next, we relax the isolated feature importance assumption. To incorporate feature interdependencies in the interpretation,  we define the importance function over subsets of features, referred to as {\it group-features}. We adjust the subset size on a per-example basis using an unsupervised approach, making the interpretation {\it context-aware}. Including group-features in the interpretation makes the optimization combinatorial. To circumvent the associated computational issues, we use an $L_0-L_1$ relaxation as is common in compressive sensing~\cite{candes2005decoding,donoho2006compressed}, LASSO regression~\cite{tibshirani1996lasso}, and other related problems. To solve the relaxed optimization, we employ proximal gradient descent~\cite{Parikh:2014:PA:2693612.2693613}. Our empirical results on ImageNet indicate that incorporating group-features removes noise and makes the interpretation more visually coherent with the object of interest. We refer to our interpretation method based on first-order (gradient) information as the CAFO (Context-Aware First Order) interpretation. Similarly, the method based on second-order information is called the CASO (Context-Aware Second Order) interpretation. We provide open-source code.\footnote{\url{https://github.com/singlasahil14/CASO}}

\section{Problem Setup and Notation}\label{sec:problem_setup}

Consider a prediction problem from input variables (features) $X\in \cX \subset \mathbf{R}^d$ to an output variable $Y\in\cY$. For example, in the image classification problem, $\cX$ is the space of images and $\cY$ is the set of labels $\{1,...,c\}$. We observe $m$ samples from these variables, namely $S=\{(\bx_1,y_1),...,(\bx_m,y_m)\}$. Let $\PP_{X,Y}$ be the observed empirical distribution.\footnote{Note that for simplicity, we hide the dependency of $\PP_{X,Y}$ on $m$.} The empirical risk minimization (ERM) approach computes the optimal predictor $f_{\theta}:\cX\to \cY$ for a loss function $\ell(\cdot,\cdot)$ using the following optimization:
\begin{align}\label{opt:erm}
\min_{\theta \in \Theta}~~ \EE_{\PP_{X,Y}}\left[\ell\left(f_{\theta}(\bx),y\right)\right].
\end{align}

Let $\cS$ be a subset of $[d]:=\{1,2,...,d\}$ with cardinality $|\cS|$. For a given sample $(\bx,y)$, let $\bx(\cS)$ indicate the features of $\bx$ in positions $\cS$. We refer to $\bx(\cS)$ as a group-feature of $\bx$. The importance of a group-feature $\bx(\cS)$ is proportional to the change in the loss function when $\bx(\cS)$ is perturbed. We select the group-feature with maximum importance and visualize that subset in a saliency map.

 \begin{definition}[Group-Feature Importance Function]\label{def:group-feature-influence}
Let $\theta^*$ be the optimizer of the ERM problem \eqref{opt:erm}. For a given sample $(\bx,y)$, we define the group-feature importance function $I_{\theta^*}^{k,\rho}(\bx,y)$ as follows:
\begin{align}\label{opt:group-feature-influence}
I_{\theta^*}^{k,\rho}(\bx,y):= \max_{\tbx}~~ &\ell\left(f_{\theta^*}(\tbx),y \right)\\
&\|\tbx-\bx\|_{0} \leq k, \nonumber \\
&\|\tbx-\bx\|_{2} \leq \rho, \nonumber
\end{align}
where $\|.\|_0$ counts the number of non-zero elements of its argument (known as the $L_0$ norm). The parameter $k$ characterizes an upper bound on the cardinality of the group-features. The parameter $\rho$ characterizes an upper bound on the $L_2$ norm of feature perturbations. 
\end{definition}
If $\tbx^*$ is the solution of optimization \eqref{opt:group-feature-influence}, then the vector $|\tbx^*-\bx|$ contains the feature importance values that are visualized in the saliency map. Note, when $k=1$ this definition simplifies to current feature importance formulations which consider features in isolation. When $k>1$, our formulation can capture feature interdependencies. Parameters $k$ and $\rho$ in general depend on the test sample $\bx$ (i.e., the size of the group-features are different for each image and model). We introduce an unsupervised metric to determine these parameters in Section~\ref{subsec:tuning}, but assume these parameters are given for the time being. 

The cardinality constraint $\|\tbx-\bx\|_{0} \leq k$ (i.e. the constraint on the group-feature size) leads to a combinatorial optimization problem in general. Such a sparsity constraint has appeared in different problems such as compressive sensing \cite{candes2005decoding,donoho2006compressed} and LASSO regression \cite{tibshirani1996lasso}. Under certain conditions, we show that without loss of generality the $L_0$ norm can be relaxed with the (convex) $L_1$ norm (Appendix~\ref{sec:relaxation}).

Our goal is to solve optimization \eqref{opt:group-feature-influence} which is non-linear and non-concave in $\tbx$. Current approaches do not consider the cardinality constraint and optimize $\tbx$ by linearizing the objective function (i.e., using the gradient). 
To incorporate group-features into current methods, we can add the constraints of optimization \eqref{opt:group-feature-influence} to the objective function using Lagrange multipliers. 
This yields the following Context-Aware First-Order (CAFO) interpretation function. 

\begin{definition}[The CAFO{} Interpretation]\label{def:CAFO-feature-importance}
	For a given sample $(\bx,y)$, we define the Context-Aware First-Order (CAFO) importance function $\tI_{\theta^*}^{\lambda_1,\lambda_2}(\bx,y)$ as follows:
	\begin{equation}\label{opt:CAFO-importance}
	\begin{split}
	\tI_{\theta^*}^{\lambda_1,\lambda_2}(\bx,y):= \max_{\Delta}~~  &\grad_{\bx} \ell\left(f_{\theta^*}(\bx),y \right)^t \Delta\\ &-\lambda_1 \|\Delta\|_{1} -\lambda_2 \|\Delta\|_{2}^2 
	\end{split}
	\end{equation}
	where $\lambda_1$ and $\lambda_2$ are non-negative regularization parameters. We refer to the objective of this optimization as $\tell(\Delta)$, hiding its dependency on $(\bx,y)$ and $\theta^*$ to simplify notation. 
\end{definition}
Large values of regularization parameters $\lambda_1$ and $\lambda_2$ in optimization \eqref{opt:CAFO-importance} correspond to small values of parameters $k$ and $\rho$ in optimization \eqref{opt:group-feature-influence}. Incorporating group-features naturally leads to a sparsity regularizer through the $L_1$ penalty. Note, this is not a hard constraint which forces a sparse interpretation. Instead, given proper choice of the regularization coefficients, the interpretation will reflect the sparsity used by the underlying model. In Section~\ref{subsec:tuning}, we detail our method for setting $\lambda_1$ on an example-specific basis (i.e., context-aware) based on the sparsity ratio of CAFO's optimal solution. Moreover, in Appendix~\ref{sec:relaxation}, we show that under some general conditions, optimization \eqref{opt:CAFO-importance} can be solved efficiently and its solution matches that of the original optimization \eqref{opt:group-feature-influence}. 

To better approximate the loss function, we use its second-order Taylor expansion around point $(\bx,y)$:
\begin{equation}\label{eq:taylor}
\begin{split}
\ell\left(f_{\theta^*}(\tbx),y \right) \approx & \ell\left(f_{\theta^*}(\bx),y \right)+ \underbrace{\grad_{\bx} \ell\left(f_{\theta^*}(\bx),y \right)^t \Delta}_{\text{the first-order term}} \\&+ \underbrace{\frac{1}{2} \Delta^t \bH_{\bx} \Delta}_{\text{the second-order term}}
\end{split} 
\end{equation}
where $\Delta:=\tbx-\bx$ and $\bH_{\bx}$ is the Hessian of the loss function on the input features $\bx$ (note $y$ is fixed). This second-order expansion of the loss function decreases the interpretation's model approximation error. 

By choosing proper values for regularization parameters, the resulting optimization using the second-order surrogate loss is strictly a convex minimization (or equivalently concave maximization) problem, allowing for efficient optimization using gradient descent (Theorem \ref{thm:concave}). Moreover, even though the Hessian matrix $\bH_{\bx}$ can be expensive to compute for large neural networks, gradient updates of our method only require the Hessian-vector product (i.e., $\bH_{\bx} \Delta$) which can be computed efficiently \cite{pearlmutter1994hessian}. This yields the following Context-Aware Second-Order (CASO) interpretation function.

\begin{definition}[The \name{} Interpretation]\label{def:CASO-feature-importance}
For a given sample $(\bx,y)$, we define the Context-Aware Second-Order (CASO) importance function $\tI_{\theta^*}^{\lambda_1,\lambda_2}(\bx,y)$ as follows:
\begin{equation}\label{opt:CASO-importance}
\begin{split}
\tI_{\theta^*}^{\lambda_1,\lambda_2}(\bx,y):= \max_{\Delta}~~  &\grad_{\bx} \ell\left(f_{\theta^*}(\bx),y \right)^t \Delta +\frac{1}{2} \Delta^t \bH_{\bx} \Delta\\
&-\lambda_1 \|\Delta\|_{1} -\lambda_2 \|\Delta\|_{2}^2 
\end{split}
\end{equation}
We refer to the objective of this optimization as $\tell(\Delta)$. $\lambda_1$ and $\lambda_2$ are defined as in \eqref{opt:CAFO-importance}. 
\end{definition}

\section{The Impact of the Hessian}\label{sec:hessian_impact}
The Hessian is by definition useful when the loss function at the test sample has high curvature. However, given the linear nature of popular network architectures with piecewise linear activations, e.g., ReLU~\cite{glorot2011relu} or Maxout~\cite{goodfellow2013maxout}, \textit{do these regions of high curvature even exist?} We answer this question for neural networks with piecewise linear activations by first providing an exact calculation of the input Hessian. Then, we use this derivation to understand the impact of including the Hessian term in interpretation. More specifically, we prove that when the probability of the predicted class is $\approx$ 1 and the number of classes is large, the second-order interpretation is similar to the first-order one. We verify this theoretical result experimentally over images in the ImageNet 2012 dataset~\cite{russakovsky2015imagenet}. We also observe that when the confidence in the predicted class is low, the second-order interpretation can be significantly different from the first-order interpretation. Since second-order interpretations take into account the curvature of the model, we conjecture that they are more faithful to the underlying model in these cases.

\subsection{Closed-form Hessian Formula for ReLU Networks}\label{subsec:hessian-decompose-intro}

We present an abridged version of the exact Hessian calculation here, the details are provided in Appendix~\ref{proof:hessian_formula}. Neural network models which use piecewise linear activation functions have class scores (logits) which are linear functions of the input. That is, since they are piecewise linear over the entire domain, they are linear at a particular input.\footnote{Note that we ignore points where the function is non-differentiable as they form a measure zero set.}
Thus, we can write:
\begin{align*}
f_{\theta}(\bx) &=  \bWW^T\bx + \bb,
\end{align*}
where $\bx$ is the input of dimension $d$, $f_{\theta}(\bx)$ are the logits, $\bWW$ are the weights, and $\bb$ are the biases of the linear function. Note that $\bWW$ combines weights of different layers from the input to the output of the network. Each row $\bWW_{i}$ of $\bWW$ is the gradient of logit $f_{\theta}(\bx)_{i}$ with respect to the flattened input $\bx$ and can be handled in auto-grad software such as PyTorch~\cite{paszke2017pytorch}. We define:
\begin{align*}
&\bp = \text{softmax}(f_{\theta}(\bx))\\
&\ell(f_{\theta}(\bx),y) = -\sum_{i=1}^{\text{c}} y_i \text{log}(\bp_i),
\end{align*}
where $c$ denotes the number of classes, $\bp$ denotes the class probabilities, and $\ell(\bp,\by)$ is the cross-entropy loss function.

 In this case, we have the following result:
\begin{proposition}\label{prop:hessian-formula}
$\bH_{\bx}$ is given by:
\begin{align}\label{eq:hessian-formula}
\bH_{\bx} = \nabla^2_{\bx} {\ell(\bp,\by)} &= \bWW(\text{diag}(\bp) - \bp\bp^T)\bWW^T
\end{align}
\end{proposition}
where $\text{diag}(\bp)$ is a diagonal matrix whose diagonal elements are equal to $\bp$. 

The first observation from Proposition \ref{prop:hessian-formula} is as follows:
\begin{theorem}\label{thm:A_psd}
	$\bH_{\bx}$ is a positive semidefinite matrix.
\end{theorem}
These two results allow an extremely efficient computation of the Hessian's eigenvectors and eigenvalues using the Cholesky decomposition of $\text{diag}(\bp) - \bp\bp^T$ (Appendix~\ref{sec:decomposition}). Note the use of decomposition is critical as storing the Hessian requires intractable amounts of memory for high dimensional inputs. The entire calculation of the Hessian's decomposition for ImageNet using a ResNet-50~\cite{he2016resnet} runs in approximately 4.2 seconds on an NVIDIA GTX 1080 Ti. 

To the best of our knowledge, this is the first work which derives the exact Hessian decomposition for piecewise linear networks. \citet{yao2018hessian} also proved the Hessian for piecewise linear networks is at most rank $c$ but did not derive the exact input Hessian.

One advantage of having a closed-form formula for the Hessian matrix \eqref{eq:hessian-formula} is that we can use it to properly set the regularization parameter $\lambda_2$ in CASO's formulation. To do this, we rely on the following result:

\begin{theorem}\label{thm:concave}
	If $L$ is the largest eigenvalue of $\bH_{\bx}$, for any value of $\lambda_2>L/2$, the second-order interpretation objective function \eqref{opt:CASO-importance} is strongly concave.
\end{theorem}

We use Theorem~\ref{thm:concave} to set the regularization parameter $\lambda_2$ for CASO. We need to set $\lambda_2$ to make the optimization convex, but not set $\lambda_2$ so large that it overpowers $\bH_{\bx}$. In particular, we set $\lambda_2 = L/2 + c_1$, where we choose $c_1=10$ for CASO and CAFO. We observe that if $c_1$ is small, the optimization can become non-convex due to numerical error in the calculation of L. However above a threshold, the value of $c$ does not have a significant impact on the saliency map. 

\subsection{Theoretical Results on the Hessian Impact}\label{subsec:l2_results}

We now leverage the exact Hessian calculation to prove that when the probability of predicted class is $\approx$ 1 and the number of classes is large, the Hessian of a piece-wise linear neural network is approximately of rank one and its eigenvector is approximately parallel to the gradient. Since a constant scaling does not affect the visualization, this causes the two interpretations to be similar to one another. 

\begin{theorem}\label{thm:hessianparallel}
	If the probability of the predicted class=1-(c-1)$\eps$ , where $\eps\approx0$, then as c $\to \infty$ such that $c\eps \approx 0$, Hessian is of rank one and its eigenvector is parallel to the gradient.
\end{theorem}
Let $\Delta_{\name{}}^*$ be the optimal solution to the CASO objective \ref{opt:CASO-importance} and $\Delta_{\firstorder{}}^*$ be the optimal solution for the CAFO objective \ref{opt:CAFO-importance}. We assume $\lambda_1$=0 for both the objectives. 

\begin{theorem}\label{thm:equivalence}
	If the probability of the predicted class=1-(c-1)$\eps$ , where $\eps\approx0$, then as c $\to \infty$ such that $c\eps \approx 0$, the CASO solution \eqref{opt:CASO-importance} with $\lambda_1=0$ is almost parallel to the CAFO solution \eqref{opt:CAFO-importance} with $\lambda_1=0$.
\end{theorem}
We emphasize that our theoretical results are valid in the ``asymptotic regime''. To analyze the approximation in the finite length regime, we simulate the relative error between the true Hessian and the rank-one approximation of the Hessian as the number of classes increases and probability of predicted class tends to 1. We find the Hessian quickly converges to rank-one empirically (Appendix~\ref{subsec:sm_hessian_analysis}).

\subsection{Empirical Results on the Hessian Impact}\label{subsec:hessian_impact}

We now present empirical results on the impact of the Hessian in interpreting deep learning models. In our experiments here, we isolate the impact of the Hessian term by setting $\lambda_1=0$ in both CASO and CAFO.

A consequence of Theorem \ref{thm:concave} is that the gradient descent method with Nesterov momentum converges to the global optimizer of the second-order interpretation objective with a convergence rate of $\cO(1/t^2)$, see Appendix~\ref{sec:corollary} for details.

To optimize $\Delta$, the gradient is given by:
\begin{align}
\grad_{\Delta} \tilde{\ell}(\Delta)= \grad_{\bx} \ell\left(f_{\theta^*}(\bx),y \right) + \bH_\bx \Delta - 2\lambda_2 \Delta.
\end{align}

\begin{figure}[t]
\vskip 0.2in
\begin{center}
\centerline{\includegraphics[width=\columnwidth]{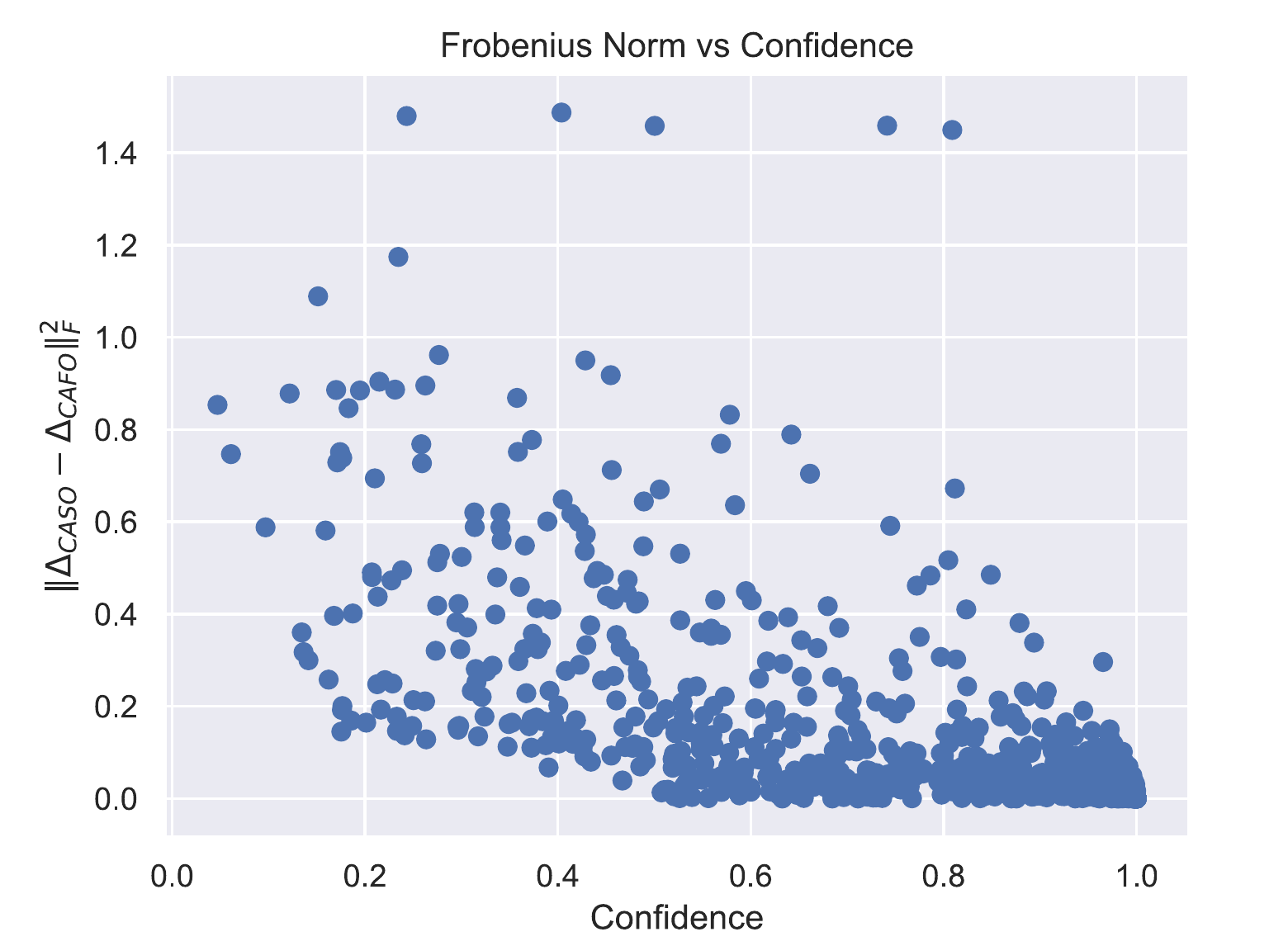}}
\caption{The Frobenius norm difference between CASO and CAFO after normalizing both vectors to have the same $L_2$ norm. Consistent with the result of Theorem \ref{thm:hessianparallel}, when the classification confidence is low, the CASO result differs significantly from CAFO. When the confidence is high, CASO and CAFO are approximately the same. To isolate the impact of the Hessian term, we assume $\lambda_1=0$ in both CASO and CAFO.}
\label{fig:scatter-hessian}
\end{center}
\vskip -0.2in
\end{figure}

\begin{figure}[ht]
\vskip 0.2in
\begin{center}
\centerline{\includegraphics[width=\columnwidth]{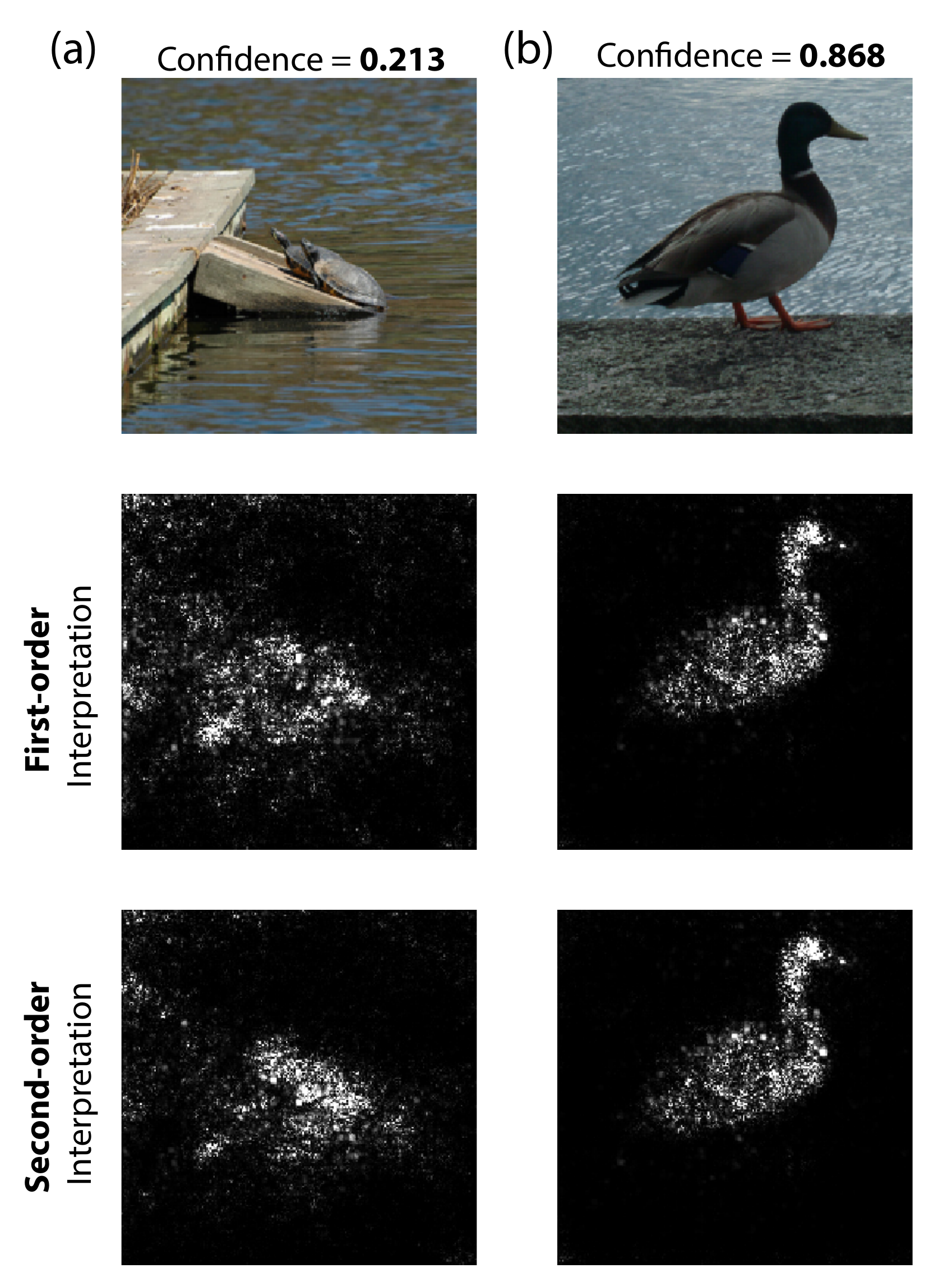}}
\caption{ Panel (a) shows an example where the classification confidence is low. In this case, the CASO and CAFO interpretations differ significantly. Panel (b) demonstrates an example where the classification confidence is high. In this case, CASO and CAFO lead to similar interpretations as suggested by our theory. }
\label{fig:scatter-hessian-ex}
\end{center}
\vskip -0.2in
\end{figure}

The gradient term $\grad_{\bx} \ell\left(f_{\theta^*}(\bx),y \right)$ and the regularization term $-2\lambda_2 \Delta$ are straightforward to implement using standard backpropagation. 

To compute the Hessian-vector product term $\bH_\bx \Delta$, we rely on the result of Pearlmutter 1994 \cite{pearlmutter1994hessian}: a Hessian-vector product can be computed in the same time as the gradient $\grad_{\bx} \ell\left(f_{\theta^*}(\bx),y \right)$. This is handled easily in modern auto-grad software. Moreover, for ReLU networks, our closed-form formula for the Hessian term (Theorem \ref{prop:hessian-formula}) can be used in the computation of the Hessian-vector product as well. In our experiments here we use the closed-form formula for $\lambda_1=0$. When $\lambda_1>0$, we use proximal gradient descent (Section~\ref{sec:group-feature}).

We compare second-order (\name{} with $\lambda_1 = 0$) and the first-order interpretations (\firstorder{} with $\lambda_1 = 0$) empirically. Note that when $\lambda_1 = 0$, $\Delta_{\firstorder{}} = \frac{1}{\lambda_2}\bg_{\bx}$ where $\bg_{\bx}$ is the gradient and $\Delta_{\firstorder{}}$ is the interpretation obtained using the \firstorder{} objective.

We compute second-order and first-order interpretations for 1000 random samples on the ImageNet ILSVRC-2012~\cite{russakovsky2015imagenet} validation set using a Resnet-50~\cite{he2016resnet} model. Our loss function $\ell(\cdot,\cdot)$ is the cross-entropy loss. After calculating $\Delta$ for all methods, the values must be normalized for visualization in a saliency map. We apply a normalization technique from existing work which we describe in Appendix~\ref{sec:sm_normalization}.

We plot the Frobenius norm of the difference between \name{} and \firstorder{} in Figure~\ref{fig:scatter-hessian}. Before taking the difference, we  normalize the $\Delta$ solutions produced by \name{} and \firstorder{} to have the same $L_2$ norm because a constant scaling of elements of $\Delta$ does not change the visualization. 

The empirical results are consistent with our theoretical results: second-order and first-order interpretations are similar when the classification confidence is high. However, when the confidence is small, including the Hessian term can be useful in deep learning interpretation. 

To observe the difference between CAFO and CASO interpretations qualitatively, we compare them for an image when the confidence is high and for one where it is low in Figure \ref{fig:scatter-hessian-ex}. When the classification confidence is high, \firstorder{} $\approx$ \name{} and when this is low, \name{} $\ne$ \firstorder{}. Additional examples have been given in Appendix~\ref{sec:sm_hessian_relu_networks}.

We do additional experiments to evaluate the impact of the Hessian on a neural network that is not piecewise linear. We interpret a SE-Resnet-50 \cite{Hu_2018_CVPR} neural network (which uses sigmoid non-linearities) on the same 1000 images. We observe a similar trend as in the case of ReLU networks (Appendix~\ref{subsec:nonrelu_networks}). 

\section{The Impact of Group-features}\label{sec:group-feature}
This section studies the impact of the group-features in deep learning interpretation. The group-feature has been included as the sparsity constraint in optimization \eqref{opt:group-feature-influence}.

To obtain an unconstrained concave optimization for the \name{} interpretation, we relaxed the sparsity (cardinality) constraint $\|\Delta\|_{0} \leq k$ (often called an $L_0$ norm constraint) to a convex $L_1$ norm constraint. Such a $L_0-L_1$ relaxation is a core component for popular learning methods such as compressive sensing~\cite{candes2005decoding,donoho2006compressed} or LASSO regression~\cite{tibshirani1996lasso}. Using results from this literature, we show this relaxation is tight under certain conditions on the Hessian matrix $\bH_{\bx}$ (see Appendix~\ref{sec:relaxation}). In other words, the optimal $\Delta$ of optimization \eqref{opt:CASO-importance} is sparse with the proper choice of regularization parameters.

Note that the regularization term $-\lambda_1 \|\Delta\|_1$ is a concave function for $\lambda_1>0$. Similarly due to Theorem~\ref{thm:concave}, the \name{} interpretation objective~\eqref{opt:CASO-importance} is strongly concave.

One method for optimizing this objective is using gradient descent as done in the second-order interpretation but using an $L_1$ regularization penalty. However, we found that this procedure leads to poor convergence properties in practice, partially due to the non-smoothness of the $L_1$ term.

To resolve this issue, we instead use \emph{proximal} gradient descent to compute a solution for CAFO and CASO when $\lambda_1> 0$. Using the Nesterov momentum method and backtracking with proximal gradient descent gives a convergence rate of $\cO(1/t^2)$ where $t$ is the number of gradient updates (Appendix~\ref{sec:corollary}). 

Below we explain how we use proximal gradient descent to optimize our objective. First, we write the objective function as the sum of a smooth and non-smooth function:
\begin{align*}
\tilde{\ell}(\Delta) = &\underbrace{\grad_{\bx} \ell\left(f_{\theta^*}(\bx),y \right)^t \Delta +\frac{1}{2} \Delta^t \bH_{\bx} \Delta- \lambda_2 \|\Delta\|_{2}^2 }_{\text{Smooth Part}} \\
&\underbrace{- \lambda_1 \|\Delta\|_{1}}_{\text{Non-Smooth Part}}\nonumber
\end{align*} 
Let $g(\Delta)$ be the smooth, $h(\Delta)$ be the non-smooth part:
$$\tilde{\ell}(\Delta) = g(\Delta) + h(\Delta)$$
\begin{align*}
&g(\Delta) = \grad_{\bx}\ell\left(f_{\theta^*}(\bx),y \right)^t \Delta +\frac{1}{2} \Delta^t \bH_{\bx} \Delta- \lambda_2 \|\Delta\|_{2}^2 \\
&h(\Delta) = - \lambda_1 \|\Delta\|_{1}
\end{align*} 
The gradient of the smooth objective is given by:
\begin{align*}
\grad_{\Delta} g(\Delta)= \grad_{\bx} \ell\left(f_{\theta^*}(\bx),y \right) + \bH_\bx \Delta - 2\lambda_2 \Delta
\end{align*}
The proximal operator is given by:
\begin{equation*}
\begin{split}
\text{prox}_{\alpha}(x) &= \argmin_{z} \frac{1}{\alpha} \|x-z\|_{2}^2 + \lambda_1\|z\|_{1} \\&= 
\begin{cases} 
x + \lambda_1\alpha & x \leq - \lambda_1\alpha \\
0       & -\lambda_1\alpha < x \leq \lambda_1\alpha\\
x - \lambda_1\alpha & \lambda_1\alpha < x
\end{cases}
\end{split}
\end{equation*}

\begin{figure*}[ht!]
	\centering
	\subfloat[Interpretation solutions for CAFO with different values of the regularization parameter $\lambda_1$.]{\includegraphics[width=\textwidth, trim={6cm 0.75cm 6cm 0.7cm}, clip]{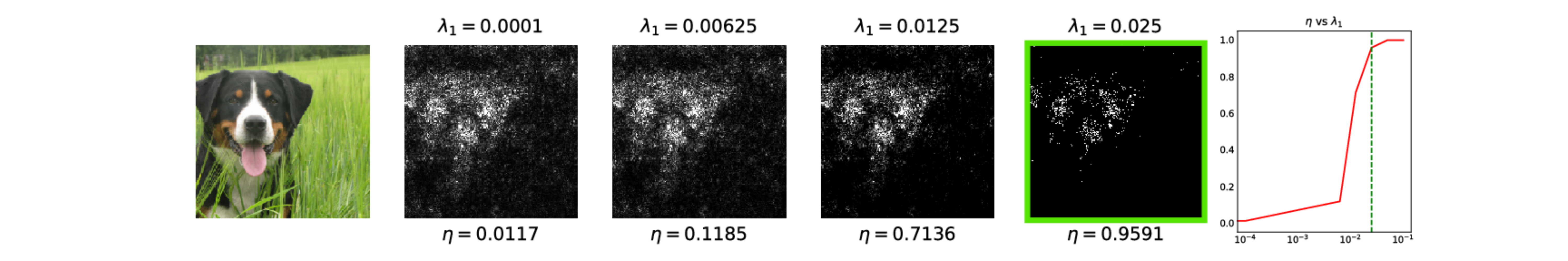}\label{fig:cafo_lambda1}}\linebreak
	\subfloat[Interpretation solutions for CASO with different values of the regularization parameter $\lambda_1$.]{\includegraphics[width=\textwidth, trim={6cm 0.75cm 6cm 0.7cm}, clip]{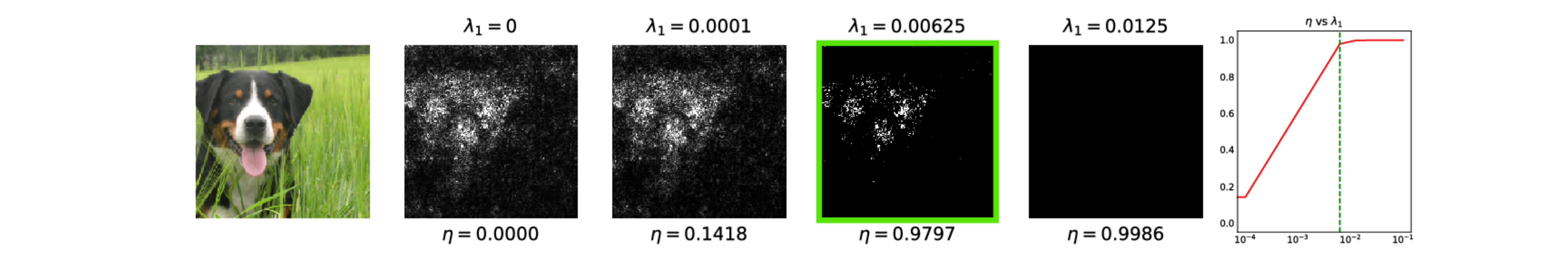}\label{fig:caso_lambda1}}\linebreak
	\subfloat[Interpretation solutions using the gradient with different clipping thresholds to induce the given sparsity.]{\includegraphics[width=\textwidth, trim={6cm 0.75cm 6cm 0.7cm}, clip]{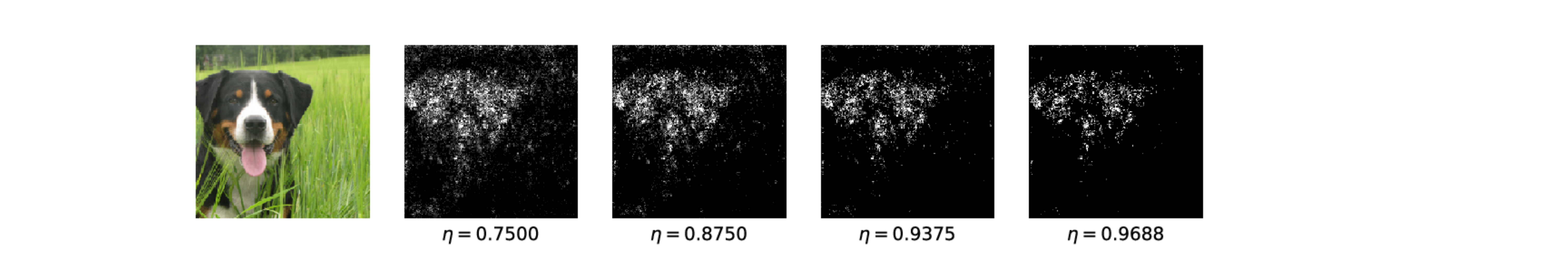}\label{fig:grad_thres}}
	\caption{Larger $\lambda_1$ values lead to higher sparsity ratios ($\eta$). Our unsupervised method selects the interpretations marked with a green box. Interpretations selected in panel (a) and (b) are less noisy compared to (c). }
	\label{fig:lambda1_sweep}
\end{figure*}

This formula can be understood intuitively as follows. If the magnitude of some elements of $\Delta$ is below a certain threshold ($\lambda_1\alpha$), proximal mapping sets those values to zero. This leads to values that are exactly zero in the saliency map.

To optimize $\Delta$, we use FISTA \cite{beck2009fista} with backtracking and the Nesterov momentum optimizer with a learning rate of $0.1$ for 10 iterations and decay factor of $0.5$. $\Delta$ is initialized to zero. FISTA takes a step with learning rate $\alpha$ to reduce the smooth objective loss $g(\Delta)$ and then applies a proximal mapping to the resulting $\Delta$. Backtracking reduces the learning rate when the update results in a higher loss.

\subsection{Empirical Impact of Group-Features}\label{subsec:tuning}

We now investigate the empirical impact of group-features. In our experiments, we focus on image classification because visual interpretations are intuitive and allow for comparison with prior work. We use a Resnet-50~\cite{he2016resnet} model on the ImageNet ILSVRC-2012 dataset. 

To gain an intuition for the effect of $\lambda_1$, we show a sweep over values in Figure~\ref{fig:lambda1_sweep}. When $\lambda_1$ is too high, the saliency map becomes all zero. Different approaches to set the regularization parameter $\lambda_1$ have been explored in different problems. For example, in LASSO, one common approach is to use Least Angle Regression \cite{2004math......6456E}. 

We propose an unsupervised method based on the sparsity ratio of the interpretation solution to set $\lambda_1$. We define $\eta$, the sparsity ratio, as the number of zero pixels divided by the total number of pixels. We start with $\lambda_1=10^{-5}$ and increase $\lambda_1$ by a factor of 10 until $\Delta$ reaches all zeros. For interpretations with sparsity in a certain range (e.g. $1 > \eta \geq 0.75$ in our examples), we choose the interpretation with the highest loss. If we do not find any interpretation that satisfies the sparsity condition, we reduce the first $\lambda_1$ that resulted in $\Delta$ becoming zero by a factor of 2 and repeat further iterations. In practice, we batch different values of $\lambda_1$ to find a reasonable parameter setting efficiently.

This method selects the interpretation marked with a green box in Figures~\ref{fig:cafo_lambda1} and \ref{fig:caso_lambda1}. In Figure~\ref{fig:grad_thres}, we show the gradient interpretation with different values of clipping thresholds to induce the specified sparsity value. We observe that the interpretations obtained using group-features (Figures~\ref{fig:cafo_lambda1} and \ref{fig:caso_lambda1}) are less noisy compared to Figure~\ref{fig:grad_thres}.

\begin{figure*}[ht!]
	\begin{center}
		\centerline{\includegraphics[width=\textwidth, trim={0cm 0cm 0cm 0cm}, clip]{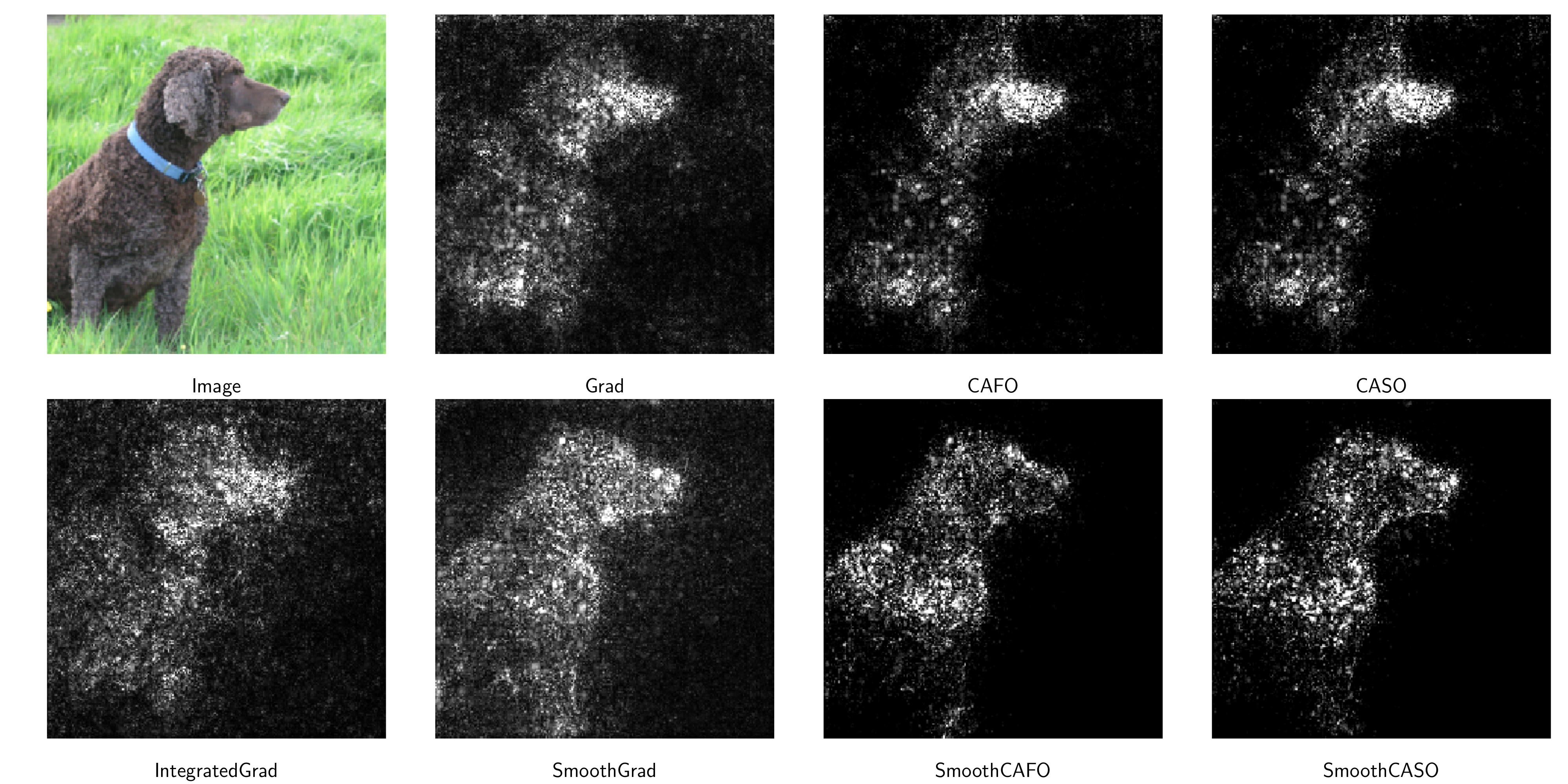}}
		\caption{A qualitative comparison of existing interpretation methods. More examples are shown in Appendix~\ref{sec:SM_existing_comp}. Grad stands for Vanilla Gradient and IntegratedGrad stands for Integrated Gradient. For our methods (CAFO, CASO, SmoothCAFO, SmoothCASO) the saliency map is more visually coherent with the object of interest compared to existing methods.}
		\label{fig:comp_existing}
	\end{center}
\end{figure*}

\section{Qualitative Comparision of Deep Learning Interpretation Methods}\label{sec:qualitative}
This section briefly reviews prior saliency map approaches and compares their performance to CAFO and CASO qualitatively. The proposed Hessian and group-feature terms can be included in existing approaches as well.

\textbf{Vanilla Gradient:} \citet{simonyan2013deep} propose to compute the gradient of the class score with respect to the input.

\textbf{SmoothGrad:} \citet{smilkov2017smoothgrad} argue that the input gradient may fluctuate sharply in the region local to the test sample. To address this, they average the gradient-based importance values generated from many noisy inputs.

\textbf{Integrated Gradients:} \citet{sundararajan2017axiomatic} define a baseline,
which represents an input absent of information (e.g., a completely zero image). 
Feature importance is determined by accumulating gradient information along the path from the baseline to the
original input: $(\bx- \bx^{\prime})\times\int_{\alpha=0}^{1} \grad_{\bx} \ell\left(f_{\theta^*}(\bx^{\prime} + \alpha (\bx-\bx^{\prime})),y \right)~d\alpha$. The integral is approximated by a finite sum.

We use the normalization method from SmoothGrad~\cite{smilkov2017smoothgrad} for visualizing the saliency map. Details of this method are given in Appendix~\ref{sec:sm_normalization}.

We can also extend the idea of SmoothGrad to define smooth versions of \name{} and \firstorder{}. This yields the following interpretation objective.

\begin{definition}[The Smooth \name{} Interpretation]\label{def:SmoothCASO-feature-importance}
	For a given sample $(\bx,y)$, we define the smooth context-aware second-order (the Smooth \name{}) importance function $\tI_{\theta^*}^{\lambda_1,\lambda_2}(\bx,y)$ as follows:
	\begin{equation}\label{opt:SmoothCASO-feature-importance}
	\begin{split}
	\tI_{\theta^*}^{\lambda_1,\lambda_2}(\bx,y):= \max_{\Delta}~~  &\frac{1}{n}\sum\limits_{1}^{n} (\grad_{\bz} \ell\left(f_{\theta^*}(\bz),y \right)^t \Delta \\&+\frac{1}{2} \Delta^t \bH_{\bz} \Delta)-\lambda_1 \|\Delta\|_{1} -\lambda_2 \|\Delta\|_{2}^2
	\end{split}
	\end{equation}
	where $\bz = \bx + N(0, \bsigma^2I)$ and $\lambda_1$ and $\lambda_2$ are defined similarly as before. \end{definition}

In the smoothed versions, we average over $n = 50$ samples with $\bsigma = 0.15$. Smooth \firstorder{} is defined similarly without the Hessian term. 

Since quantitatively evaluating a saliency map is an open problem, we focus on two qualitative aspects. First, we inspect visual coherence, i.e., only the object of interest should be highlighted and not the background. Second, we test for discriminativity, i.e., in an image with two objects the predicted object should be highlighted.

Figure~\ref{fig:comp_existing} shows comparisons between CAFO, CASO, and other existing interpretation methods. Including group-features in the interpretation leads to a sparse saliency map, eliminating the spurious noise and creating a visually coherent saliency map. More examples have been presented in Appendix~\ref{sec:SM_existing_comp}.

\section{Conclusion and Future Work}
\label{sec:conclusion}
We have studied two aspects of the deep learning interpretation problem. First, we characterized a closed-form formula for the input Hessian matrix of a deep ReLU network. Using this, we showed that, if the confidence in the predicted class is high and the number of classes is large, first-order and second-order methods produce similar results. In the process, we also proved that the Hessian matrix is of rank one and its eigenvector is parallel to the gradient. These results can be insightful in other related problems such as adversarial examples. Second, we incorporated feature interdependencies in the interpretation using a sparsity regularization term. Adding this term significantly improves qualitative interpretation results.  

There remain many open problems in interpreting deep learning models. For instance, since saliency maps are high-dimensional, they can be sensitive to noise and adversarial perturbations~\cite{ghorbani2017interpretation}. Moreover, without proper quantitative evaluation metrics for model interpretations, the evaluation of interpretations is often qualitative and can be subjective. Finally, the theoretical impact of the Hessian term for low confidence predictions and the case when the number of classes is small remains unknown. Resolving these issues are among interesting directions for future work.

\section*{Acknowledgments}
Shi Feng and Eric Wallace were supported by NSF Grant IIS-1822494.
Any opinions, findings, conclusions, or recommendations expressed here
are those of the authors and do not necessarily reflect the view of
the sponsor.


\vspace{0.5cm}

\appendix

{\Large \bf Appendix}

\section{Proofs}     

\subsection{Proof of Proposition \ref{prop:hessian-formula} }\label{proof:hessian_formula}
This section derives the closed-form formula for the Hessian of the loss function for a deep ReLU network. Since a ReLU network is piecewise linear, it is locally linear around an input $\bx$. Thus the logits can be represented as:
\begin{align*}
f_{\theta}(\bx) &=  \bWW^T\bx + \bb,
\end{align*}
where $\bx$ is the input of dimension $d$, $f_{\theta}(\bx)$ are the logits, $\bWW$ are the weights, and $\bb$ are the biases of the linear function. In this proof, we use $\hby$ to denote the logits, $\bp$ to denote the class probabilities, $\by$ to denote the label vector and c to denote the number of classes. Each column $\bWW_{i}$ of $\bWW$ is the gradient of logit $\hby_{i}$ with respect to flattened input $\bx$ and can be easily handled in auto-grad software such as PyTorch~\cite{paszke2017pytorch}. 

Thus
\begin{align}
&\frac{\partial \hby_i}{\partial \bx} = \bWW_{i} \label{logit_input_grad}\\
&\bp = \text{softmax}(\hby) \nonumber\\
&\ell(\bp, \by) = -\sum_{i=1}^{\text{c}} \nonumber \by_i\text{log}(\bp_i).\\
&\nabla_{\hby} {\ell(\bp,\by)} = \bp-\by \nonumber \\
&\implies \frac{\partial \ell(\bp,\by)}{\partial \hby_i} = \bp_{i}-\by_{i} \label{loss_logit_grad}\\
&\nabla_{\bx} {\ell(\bp,\by)} = \sum_{i=1}^{\text{c}} \frac{\partial \hby_i}{\partial \bx} \times \frac{\partial \ell(\bp,\by)}{\partial \hby_i} \nonumber
\end{align}
Using \eqref{logit_input_grad} and \eqref{loss_logit_grad}
\begin{align}
&\nabla_{\bx} {\ell(\bp,\by)} = \sum_{i=1}^{\text{c}} \bWW_{i}(\bp_{i}-\by_{i}) \nonumber\\
&\implies \nabla_{\bx} {\ell(\bp,\by)} = \bWW(\bp-\by) \nonumber
\end{align}

Therefore, we have:
\begin{align}
\bH_{\bx} &= \nabla_{\bx} (\nabla_{\bx} {\ell(\bp,\by)}) = \nabla_{\bx} (\sum_{i=1}^{\text{c}} \bWW_{i}(\bp_{i}-\by_{i})) \nonumber\\
\bH_{\bx} &= \sum_{i=1}^{\text{c}} \bWW_{i}(\nabla_{\bx}(\bp_{i}-\by_{i}))^T \nonumber\\
\bH_{\bx} &= \sum_{i=1}^{\text{c}} \bWW_{i}(\nabla_{\bx}\bp_{i})^T \label{hessian_equation}
\end{align}
Deriving $\nabla_{\bx}\bp_{i}$:
\begin{align}
\nabla_{\bx} {\bp_{i}} &= \sum_{j=1}^{\text{c}}  \frac{\partial \hby_j}{\partial \bx} \times \frac{\partial \bp_{i}}{\partial \hby_j} \nonumber\\ 
\implies \nabla_{\bx} {\bp_{i}} &= \sum_{j=1}^{\text{c}}  \bigg(\bWW_{j} \times \frac{\partial \bp_{i}}{\partial \hby_j}\bigg) \quad (\text{Using } \eqref{logit_input_grad}) \label{prob_input_grad}\\
\frac{\partial \bp_{i}}{\partial \hby_j} &= 
\begin{dcases} 
\bp_{i} - \bp_{i}^{2}  & i = j \\
- \bp_{i}\bp_{j}      & i \ne j
\end{dcases} \nonumber\\
\implies \nabla_{\hby} {\bp} &= \text{diag}(\bp) - \bp\bp^T \label{midmatrix}
\end{align}
\begin{align*}
&\bH_{\bx} = \sum_{i=1}^{\text{c}} \bWW_{i}\bigg(\sum_{j=1}^{\text{c}} \bWW_{j} \times \frac{\partial \bp_{i}}{\partial \hby_j}\bigg)^T \quad (\text{Substituting } \eqref{prob_input_grad} \text{ in } \eqref{hessian_equation})\\ 
&\bH_{\bx} = \sum_{i=1}^{\text{c}}\sum_{j=1}^{\text{c}} \bWW_{i} \frac{\partial \bp_{i}}{\partial \hby_j} \bWW_{j}^T\\
&\implies \bH_{\bx} = \bWW(\text{diag}(\bp) - \bp\bp^T)\bWW^T \quad \text{(Using } \eqref{midmatrix})\\
\end{align*}
Thus we have,
\begin{align}
\nabla_{\bx} {\ell(\bp,\by)} &= \bg_{\bx} =  \bWW(\bp-\by)\label{eq:gradient}\\
\bH_{\bx} &= \bWW\bA\bWW^T\label{eq:hessian}
\end{align}
where 
\begin{align}\label{eq:A_mat}
\bA := \text{diag}(\bp) - \bp\bp^T.
\end{align}
This completes the proof.

\subsection{Proof of Theorem \ref{thm:A_psd}}\label{proof:A_psd}
To simplify notation, define $\bA$ as in \eqref{eq:A_mat}. For any arbitrary row of the matrix $\bA_{i}$, we have
\begin{align*}
\sum_{\text{j} \ne \text{i}} |\bA_{ij}| &= (\sum_{\text{j} \ne \text{i}}|-\bp_{i}\bp_{j}|)\\
\implies \sum_{\text{j} \ne \text{i}} |\bA_{ij}| &= \bp_{i}\sum_{\text{j} \ne \text{i}} \bp_{j}\\
\implies \sum_{\text{j} \ne \text{i}} |\bA_{ij}| &= \bp_{i}(1-\bp_{i})\\
|\bA_{ii}| &= \bp_{i}(1-\bp_{i})
\end{align*}
Because $|\bA_{ii}| >= \sum_{\text{j} \ne \text{i}} |\bA_{ij}|$, 
by the Gershgorin Circle theorem, we have that all eigenvalues of $\bA$ are positive and $\bA$ is a positive semidefinite matrix. Since $\bA$ is positive semidefinite, we can write $\bA = \bL\bL^T$. Using \eqref{eq:hessian}:
$$ \bH_{\bx} = \bWW\bA\bWW^T = \bWW\bL\bL^T\bWW^T = \bWW\bL(\bWW\bL)^T. $$
Hence $\bH_{\bx}$ is a positive semidefinite matrix as well.

\subsection{Proof of Theorem \ref{thm:concave}}\label{proof:hessian_concave}
The second-order interpretation objective function is:
\begin{align*}
\tilde{\ell}(\Delta) &= \grad_{\bx} \ell\left(f_{\theta^*}(\bx),y \right)^t \Delta + \frac{1}{2} \Delta^t \bH_{\bx} \Delta- \lambda_2 \|\Delta\|_{2}^2 \\
\tilde{\ell}(\Delta) &= \grad_{\bx} \ell\left(f_{\theta^*}(\bx),y \right)^t \Delta + \frac{1}{2} \Delta^t (\bH_{\bx} - 2\lambda_2\bI) \Delta 
\end{align*} 
where $\Delta:=\tbx-\bx$ ($y$ is fixed).  Therefore if $\lambda_2>L/2$, $\bH_{\bx} - 2\lambda_2\bI$ is negative definite and $\tilde{\ell}(\Delta)$ is strongly concave.

\subsection{Proof of Theorem \ref{thm:hessianparallel}}\label{proof:hessianparallel}

Let the class probabilities be denoted by $\bp$, the number of classes by c and the label vector by $\by$. We again use $\bg_{\bx}$ and $\bH_{\bx}$ as defined in \eqref{eq:gradient} and \eqref{eq:hessian} respectively.
Without loss of generality, assume that the first class is the class with maximum probability. Hence,
\begin{align}
\by &= [1, 0, 0,..., 0]^T.\label{label_vector}
\end{align}
We assume all other classes have small probability (i.e., the confidence is high),
$$\bp_{i} = \eps \approx 0\quad ~\forall \text{ i} \in \text{[2, c]}$$
\text{Since } $\sum_{i=1}^{c} \bp_{i} = 1$,
\begin{align}
&\implies \bp_{1} = 1-(c-1)\eps, \nonumber\\
&\implies \bp = [1-(c-1)\eps,\ \eps,\dotsc,\ \eps]^T 
\label{prob_vector}
\end{align}
We define:
$$\bA = \text{diag}(\bp) - \bp\bp^T$$ 
\begin{align*}
&\bA = \begin{bmatrix}
a_{11} &  a_{12}  & \ldots & a_{1c}\\
a_{21}  &  a_{22} & \ldots & a_{2c}\\
\vdots & \vdots & \ddots & \vdots\\
a_{c1}  &   a_{c2}       &\ldots & a_{cc}
\end{bmatrix}\\
&a_{11} = 1-(c-1)\eps -(1-(c-1)\eps)^2\\
&a_{1i} = a_{i1} = -(1-(c-1)\eps)\eps \quad \forall\ i\in [2, c]\\
&a_{ii} = \eps-\eps^2 \quad \forall\ i \in [2, c]\\
&a_{ij} = -\eps^2 \quad \forall\ i,j \in [2, c],\ i \ne j
\end{align*}
Ignoring $\eps^2$ terms:
\begin{align*}
&a_{11} = (c-1)\eps\\
&a_{1i} = a_{i1} = -\eps \quad \forall i\in [2, c]\\
&a_{ii} = \eps \quad \forall i \in [2, c]\\
&a_{ij} = 0 \quad \forall i,j \in [2, c], i \ne j
\end{align*}
Let $\lambda$ be an eigenvalue of $\bA$ and $\bv$ be an eigenvector of $\bA$, then $\bA\bv = \lambda\bv$.\\
Let $v_{1}, v_{2}, \dots, v_{n}$ be the individual components of the eigenvector. The equation $\bA\bv = \lambda\bv$ can be rewritten in terms of its individual components as follows:
\begin{align}
& (c-1)\eps v_{1} - \eps\sum_{i=2}^{c}v_{i} = \lambda v_{1}\label{v1_equation}\\
& -\eps v_{1} + \eps v_{i} = \lambda v_{i} \text{ } \forall i \in [2,c] \nonumber\\
& \implies v_{i} = \frac{\eps}{\eps-\lambda}v_{1} \text{ } \forall i \in [2, c] \text{, for } \lambda \ne \eps \label{eps_neq_lambda}\\
& \implies \text{or } v_{1} = 0 \text{, for } \lambda=\eps \label{eps_eq_lambda}
\end{align}

We first consider the case $\lambda \ne \eps\ \eqref{eps_neq_lambda}$. Substituting $v_{i}$ in $\eqref{v1_equation}$:
\begin{align}
(c-1)\eps v_{1} - \eps\sum_{i=2}^{c}v_{i} &= (c-1)\eps v_{1} - \frac{\eps^2}{\eps-\lambda}\sum_{i=2}^{c}v_{1} \nonumber\\
&= (c-1)\eps v_{1} - \frac{\eps^2}{\eps-\lambda}(c-1)v_{1} \nonumber\\
&= (c-1)\eps v_{1} - (c-1)\eps v_{1}\frac{\eps}{\eps-\lambda} \nonumber\\
&= \lambda v_{1} \nonumber
\end{align}
\begin{align}
&(c-1)\eps v_{1}\bigg[1-\frac{\eps}{\eps-\lambda}\bigg] = \lambda v_{1} \nonumber\\
&(c-1)\eps v_{1}\bigg[-\frac{\lambda}{\eps-\lambda}\bigg] = \lambda v_{1} \nonumber\\
&(c-1)\eps v_{1}(-\lambda) = \lambda v_{1}(\eps-\lambda)\nonumber\\
&\implies \lambda v_{1}(c\eps-\lambda) = 0 \nonumber\\
&\implies \lambda = 0 \text{ or } v_{1} = 0  \text{ or } \lambda = c\eps \nonumber\\
&v_{1}=0 \implies v_{i} =\frac{\eps}{\eps-\lambda} v_{1}=0 \quad \forall \ i \in [2,c] \nonumber\\
&\implies \bv=0 \nonumber
\end{align}
Since $\bv$ is an eigenvector, it cannot be zero,\\
$$\implies \lambda = 0 \text{ or } \lambda = c\eps. \nonumber$$
Let $\bu_{1}$ be the corresponding eigenvector for $\lambda = c\eps$.\\
By substituting $\lambda = c\eps$ in \eqref{eps_neq_lambda}:
\begin{align}
&\bu_{1}^T \propto [1-c, 1,..., 1] \nonumber
\end{align}
Dividing by the normalization constant,\\
\begin{align}
&\bu_{1}^T = \frac{1}{\sqrt{c(c-1)}}[1-c, 1,..., 1] \label{u1_equation}
\end{align}
Now we consider the case $\lambda=\eps\ \eqref{eps_eq_lambda}$. Substituting $v_{1}=0,\ \lambda = \eps$ in \ $\eqref{v1_equation}$: \\
The space of eigenvectors for $\lambda=\eps$ is a $c-2$ dimensional
subspace with $v_{1} = 0,\ \sum_{i=2}^{c}v_{i} = 0.$\\
Let $\bu_{i}$ be the eigenvectors with $\lambda=\eps \quad \forall $   $i \in [2, c-1]$\\
Let $\bu_{c}$ be the eigenvector with $\lambda=0.$\\
Writing $\bA$ in terms of its eigenvalues and eigenvectors,
\begin{align*}
\bA = c\eps \bu_{1}\bu_{1}^T + \eps\sum_{i=2}^{c-1} \bu_{i}\bu_{i}^T
\end{align*}
Let
$$\bA_{1} = c\eps \bu_{1}\bu_{1}^T, \quad \bA_{2} = \eps\sum_{i=2}^{c-1} \bu_{i}\bu_{i}^T$$
$$\| \bA_{1}\|_{F} = c\eps,\quad \| \bA_{2}\|_{F} = \eps\sqrt{c-2}$$
Hence as $c\to\infty$, \\
$$\bA = \bA_{1} + \bA_{2} \approx \bA_{1}$$
Using $\eqref{eq:hessian}$,
$$\bH_{\bx} = \bWW\bA\bWW^T \approx \bWW\bA_{1}\bWW^T$$
Substituting\ $\bA_{1} =c\eps \bu_{1}\bu_{1}^T$,
\begin{align}
&\bH_{\bx} \approx c\eps\bWW\bu_{1}\bu_{1}^T\bWW^T \label{hessian_approx}
\end{align}
Using  \eqref{eq:gradient}, 
$$\bg_{\bx} = \nabla_{\bx} {\ell(\bp,\by)} = \bWW(\bp-\by)$$
Let $\bWW_{i}$ denote the $i^{th}$ row of $\bWW$,\\
Using $\eqref{label_vector}$ and $\eqref{prob_vector}$, \\
$$\bg_{\bx} = \bWW_{1}(1-c)\eps + \sum_{i=2}^{c}\bWW_{i}\eps$$\\
$$\bg_{\bx} = \eps({\bWW_{1}(1-c) + \sum_{i=2}^{c}\bWW_{i}})$$
Using \eqref{u1_equation}, \\
$$\bg_{\bx} = \eps\sqrt{c(c-1)}\bWW\bu_{1}$$
\begin{align}
&\implies \bWW\bu_{1} = \frac{\bg_{\bx}}{\eps\sqrt{c(c-1)}} \label{grad_hessian_trick}
\end{align}
Using \eqref{hessian_approx},\\
$$\bH_{\bx} \approx c\eps\bWW\bu_{1}\bu_{1}^T\bWW^T = c\eps\bWW\bu_{1}(\bWW\bu_{1})^T $$
Using \eqref{grad_hessian_trick},\\
$$\bH_{\bx} \approx c\eps\frac{\bg_{\bx}}{\eps\sqrt{c(c-1)}}\frac{\bg_{\bx}^T}{\eps\sqrt{c(c-1)}} $$
$$\bH_{\bx} \approx c\eps\frac{\bg_{\bx}\bg_{\bx}^T}{\eps^2c(c-1)} = \frac{\bg_{\bx}\bg_{\bx}^T}{\eps(c-1)}$$
\begin{align}
\implies  \bH_{\bx} \approx \frac{\bg_{\bx}\bg_{\bx}^T}{\eps(c-1)} \label{hessian_approx_final}
\end{align}

Thus, the Hessian is approximately rank one and the gradient is parallel to the Hessian's only eigenvector.

\subsection{Proof of Theorem \ref{thm:equivalence}}\label{proof:equivalence}
We use $\bg_{\bx} = \nabla_{\bx} {\ell(\bp,\by)} = \bWW(\bp-\by)$ \quad \eqref{eq:gradient}.\\
Let $\lambda_1$ = 0 in the \name{} and \firstorder{} objectives. The CASO objective then becomes:
\begin{align*}
&\max_{\Delta}\ (\bg_{\bx}^t \Delta +\frac{1}{2} \Delta^t \bH_{\bx} \Delta -\lambda_2 \|\Delta\|_{2}^2) 
\end{align*}

Taking the derivative with respect to $\Delta$ and solving:\\
$$\Delta^{*}_{\name} = (2\lambda_2\bI - \bH_{\bx})^{-1}\bg_{\bx}$$\\
Similarly, for the CAFO objective we get:\\
$$\Delta^{*}_{\firstorder} = \frac{1}{2\lambda_2}\bg_{\bx}$$
Using \eqref{hessian_approx_final},\\
$$\bH_{\bx} \approx \frac{\bg_{\bx}\bg_{\bx}^T}{\eps(c-1)} = \frac{\|\bg_{\bx}\|^2}{\eps(c-1)}\frac{\bg_{\bx}\bg_{\bx}^T}{\|\bg_{\bx}\|^2}$$
Define:
$$\bmu = \frac{\|\bg_{\bx}\|^2}{\eps(c-1)}$$
Thus $\bmu$  is the eigenvalue of $\bH_{\bx}$ for the eigenvector: $$\frac{\bg_{\bx}}{\|\bg_{\bx}\|}$$
Consider the matrix $\bB = (2\lambda_2\bI - \bH_{\bx})$:\\
Let $\bz_{1},\dotsc ,\bz_{d}$ be the eigenvectors of $\bB$ where:
$$\bz_{1} = \frac{\bg_{\bx}}{\|\bg_{\bx}\|}$$
Eigenvalue for $\bz_{1} = 2\lambda_2 - \mu$\\
Eigenvalue for $\bz_{i} = 2\lambda_2 \quad \forall i \in [2, d]$\\
$$\bB = (2\lambda_2 - \mu)\bz_{1}\bz_{1}^T  + 2\lambda_2\sum_{i=2}^{i=d} \bz_{i}\bz_{i}^T$$
$$\bB^{-1} = \frac{1}{(2\lambda_2 - \mu)}\bz_{1}\bz_{1}^T  + \frac{1}{2\lambda_2}\sum_{i=2}^{i=d} \bz_{i}\bz_{i}^T$$
\begin{align*}
\bB^{-1} &= \frac{1}{(2\lambda_2 - \mu)}\frac{\bg_{\bx}\bg_{\bx}^T}{\|\bg_{\bx}\|^2}  + \frac{1}{2\lambda_2}\sum_{i=2}^{i=d} \bz_{i}\bz_{i}^T\\
\Delta^{*}_{\name} &= \bB^{-1}\bg_{\bx}\\
\Delta^{*}_{\name} &= \left[\frac{1}{(2\lambda_2-\bmu)}\frac{\bg_{\bx}\bg_{\bx}^T}{\|\bg_{\bx}\|^2} + \frac{1}{2\lambda_2}\sum_{i=2}^{i=d}\bz_{i}\bz_{i}^T\right]\bg_{\bx}
\end{align*}
Since each $\bz_{i}$ is orthogonal to $\bg_{\bx}$
\begin{align*}
&\implies \Delta^{*}_{\name} = \frac{\bg_{\bx}}{(2\lambda_2-\bmu)} = \frac{2\lambda_2 \Delta^{*}_{\firstorder}}{(2\lambda_2-\bmu)}
\end{align*}
Hence $\Delta^{*}_{\name}$ $\parallel$ $\Delta^{*}_{\firstorder}$ and since scaling does not affect the visualization, the two interpretations are equivalent.

\section{Convergence of Gradient Descent to Solve \name{}}\label{sec:corollary}
A consequence of Theorem \ref{thm:concave} is that gradient descent converges to the global optimizer of the second-order interpretation objective objective with a convergence rate of $\cO(1/t^2)$. More precisely, we have:

\begin{corollary}
	\label{corollary:convergence}
	Let $\tilde{\ell}(\Delta)$ be the objective function of the second-order interpretation objective (Definition \ref{def:CASO-feature-importance}). Let $\Delta^{(t)}$ be the value of $\Delta$ in the $t^\text{th}$ step with a learning rate $\alpha \leq \lambda_2-L/2$. We have
	\begin{align*}
	\tilde{\ell}(\Delta^{(t)})-\tilde{\ell}(\Delta^*) \leq \frac{2\|\Delta^{(0)}-\Delta^*\|_2^2}{\alpha(t+1)^2}.
	\end{align*} 
\end{corollary}

\section{Efficient Computation of the Hessian Matrix Using the Cholesky Decomposition}\label{sec:decomposition}
By Theorem \ref{thm:A_psd}, the Cholesky decomposition of $\bA$ (defined in \eqref{eq:A_mat}) exists. Let $\bL$ be the Cholesky decomposition of $\bA$. Thus, we have 
\begin{align*}
&\bA = \bL\bL^T\\
&\bH_{\bx} = \bWW\bL\bL^T\bWW^T\\
\end{align*}
Let $\bB := \bWW\bL$. Thus, $\bH_{\bx}$ can be re-written as $\bH_{\bx} = \bB\bB^T$. 

Let the SVD of $\bB$ be as the following:
\begin{align*}
\bB = \bU\bSigma\bV^T
\end{align*}
Thus, we can write:
\begin{align*}
\bH_{\bx} = \bU\bSigma^2\bU^T
\end{align*}
Define $\bC=\bB^T\bB = \bV\bSigma^2\bV$. Note that $\bSigma^2$, the eigenvalues of $\bC$ and $\bH_{\bx}$ are the same. For a dataset such as ImageNet, the input has dimension d = 224$\times$224$\times$3 and c = 1000.  Decomposing C (size 1000$\times$1000) into its eigenvalues $\bSigma$ and eigenvectors $\bV$ is computationally efficient. Thus, from $\bB = \bU\bSigma\bV^T$, we can compute the eigenvectors $\bU$ of $\bH_{\bx}$.

\section{Saliency Visualization Methods}\label{sec:sm_normalization}

\textbf{Normalizing Feature Importance Values:} After assigning importance values to each input feature, the values must be normalized for visualization in a saliency map. For fair comparison across all methods, we use the non-diverging normalization method from SmoothGrad~\cite{smilkov2017smoothgrad}. This normalization method first takes the absolute value of the importance scores and then sums across the three color channels of the image. Next, the largest importance values are capped to the value of $99^\text{th}$ percentile. Finally, the importance values are divided and clipped to enforce the range $[0,1]$. Code for the method is available.\footnote{\url{https://github.com/PAIR-code/saliency/blob/master/saliency/visualization.py}}

\textbf{Domain-Specific Post-Processing:} Gradient $\odot$ Input~\cite{shrikumar2017learning} multiplies the importance values by the raw feature values. In image tasks where the baseline is zero, Integrated Gradients~\cite{sundararajan2017axiomatic} does the same. This heuristic can visually sharpen the saliency map and has some theoretical justification: it is equivalent to the original Layerwise Relevance Propagation Technique~\cite{bach2015pixelwise} modulo a scaling factor~\cite{shrikumar2017learning,kindermans2016noise}. Additionally, if the model is linear, $y = W\bx$, multiplying the gradient by the input is equivalent to a feature's true contribution to the final class score. 

However, multiplying by the input can introduce visual artifacts not present in the importance values~\cite{smilkov2017smoothgrad}. We argue against multiplying by the input: it artificially enhances the visualization and only yields benefits in the image domain. \citet{adebayo2018sanity} argue similarly and show cases when the input term can dominate the interpretation. Moreover, multiplication by the input removes the input invariance of the interpretation regardless of the invariances of the underlying model~\cite{kindermans2017unreliability}. We observed numerous failures in existing interpretation methods when input multiplication is removed.

\section{Tightness of the $L_0-L_1$ Relaxation}
\label{sec:relaxation}

We assume the condition of Theorem \ref{thm:concave} holds, thus, the \name{} optimization is a concave maximization (equivalently a convex minimization) problem.

Note the \name{} optimization with the cardinality constraint can be re-written as follows:
\begin{align}\label{opt:hec-l0}
\min_{\Delta} ~~ &\|\by-\bA \Delta\|^2,\\
& \|\Delta\|_0 \leq k,\nonumber
\end{align}
where 
\begin{align}
\label{opt:invert_A}
\bA&:= \left(\lambda_2\bI-\frac{1}{2}\bH_{\bx}\right)^{1/2}\\
\by&:= \frac{1}{2} \bA^{-1}\grad_{\bx} \ell\left(f_{\theta^*}(\bx),y \right). 
\end{align}

Where $(.)^{1/2}$ indicates the square root of a positive definite matrix. Equation~\eqref{opt:invert_A} highlights the condition for tuning the parameter $\lambda_2$: it needs to be sufficiently large to allow inversion of $\bA$ but sufficiently small to not ``overpower'' the Hessian term. Note, we are now minimizing $\Delta$ for consistency with the compressive sensing literature. To explain the conditions under which the $L_0-L_1$ relaxation is tight, we define the following notation. For a given subset $S\subset \{1,2,...,d\}$ and constant $\alpha\geq 1$, we define the following cone:
\begin{align}
\cC(S;\alpha):=\{\Delta\in \mathbb{R}^d: \|\Delta_{S^c}\|_1 \leq \|\Delta_S\|_1\},
\end{align}
where $S^c$ is the complement of $S$. We say that the matrix $\bA$ satisfies the restricted eigenvalue (RE) \cite{raskutti2010restricted,bickel2009simultaneous} condition over $S$ with parameters $(\alpha,\gamma)\in [1,\infty) \times (0,\infty)$ if
\begin{align}\label{eq:RE}
\frac{1}{d} \|\bA \Delta\|_2^2 \geq \gamma^2 \|\Delta\|_2^2 \quad \forall \Delta \in \cC(S;\alpha).
\end{align}
If this condition is satisfied for all subsets of $S$ where $|S|=k$, we say that $\bA$ satisfies the RE condition of order $k$ with parameters $(\alpha,\gamma)$. If $\bA$ satisfies the RE condition with $\alpha\geq 3$ and $\gamma >0$, then the $L_0-L_1$ relaxation of optimization \eqref{opt:hec-l0} is tight \cite{bickel2009simultaneous}. In other words, if $\Delta^*$ is the solution of optimization \eqref{opt:hec-l0}, it is also the solution of optimization   
\begin{align}\label{opt:hec-l1}
\min_{\Delta} ~~ &\|\by-\bA \Delta\|^2, \\
& \|\Delta\|_1 \leq \|\Delta^*\|_1.\nonumber
\end{align}
The Lagrange relaxation of this optimization leads to the \name{} interpretation objective. We note that the RE condition is less severe than other optimality conditions such as the restricted isometry property~\cite{candes2007dantzig}. Although it is difficult to verify that the RE condition holds for the Hessian matrix of a deep neural network, our empirical experiments are consistent with our theory: the resulting $\Delta$ of the \name{} interpretation objective is sparse for proper choices of the regularization parameters.

\section{Empirical Analysis of the Hessian Impact}\label{sec:sm_hessian_relu_networks}

\subsection{Empirically Verifying the Hessian Approximation}\label{subsec:sm_hessian_analysis}
\begin{figure}[!ht]
	\begin{center}
		\centerline{\includegraphics[width=\columnwidth]{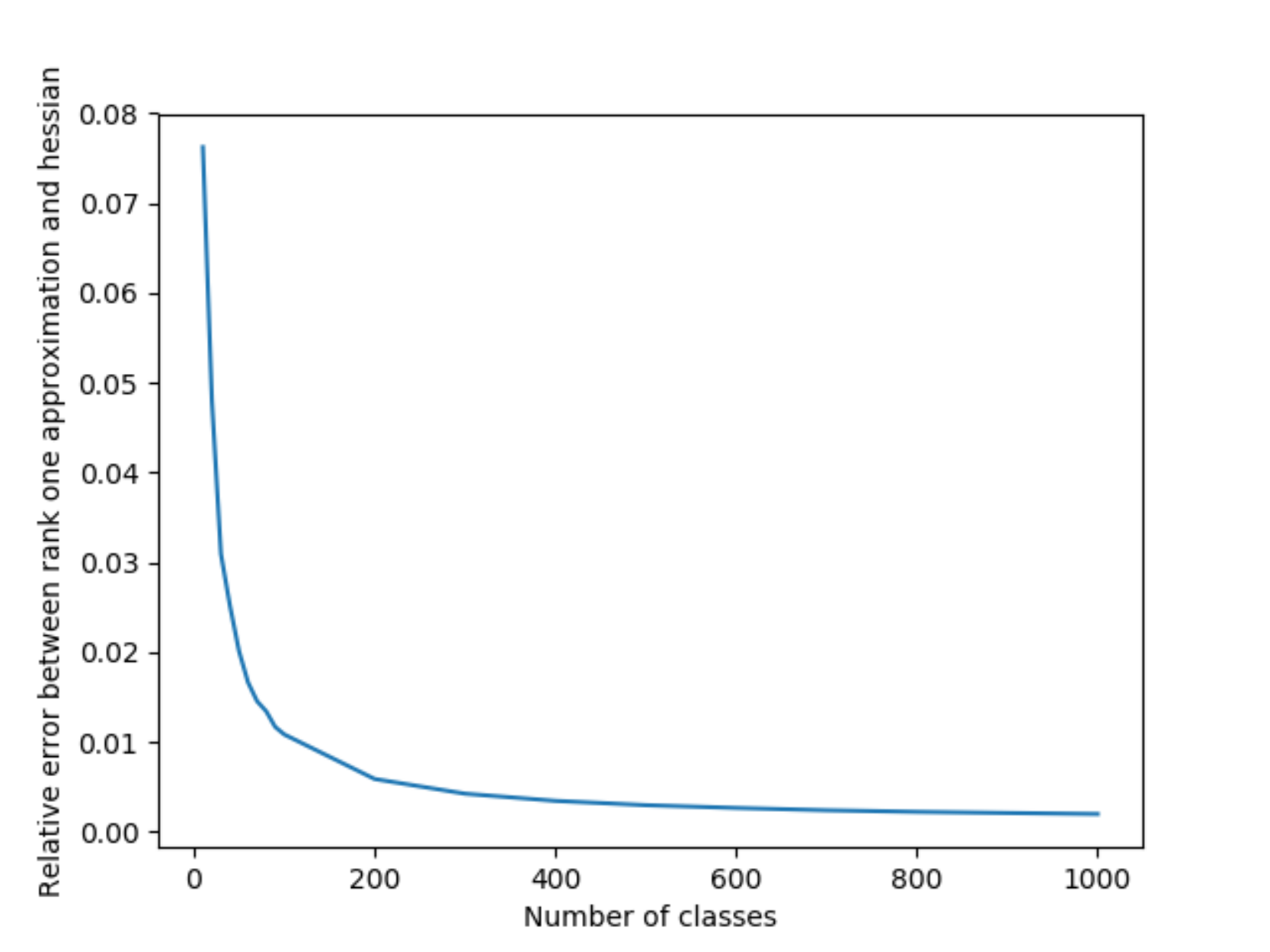}}
		\caption{The relative error between a rank one approximation of the Hessian and the true Hessian as the number of classes increases. Although our theoretical analysis only holds in the asymptotic regime, the Hessian's convergence to a rank one matrix happens quicky empirically.}
		\label{fig:rel_errors_vs_classes}
	\end{center}
\end{figure}

\begin{figure}[!ht]
	\begin{center}
		\centerline{\includegraphics[width=\columnwidth]{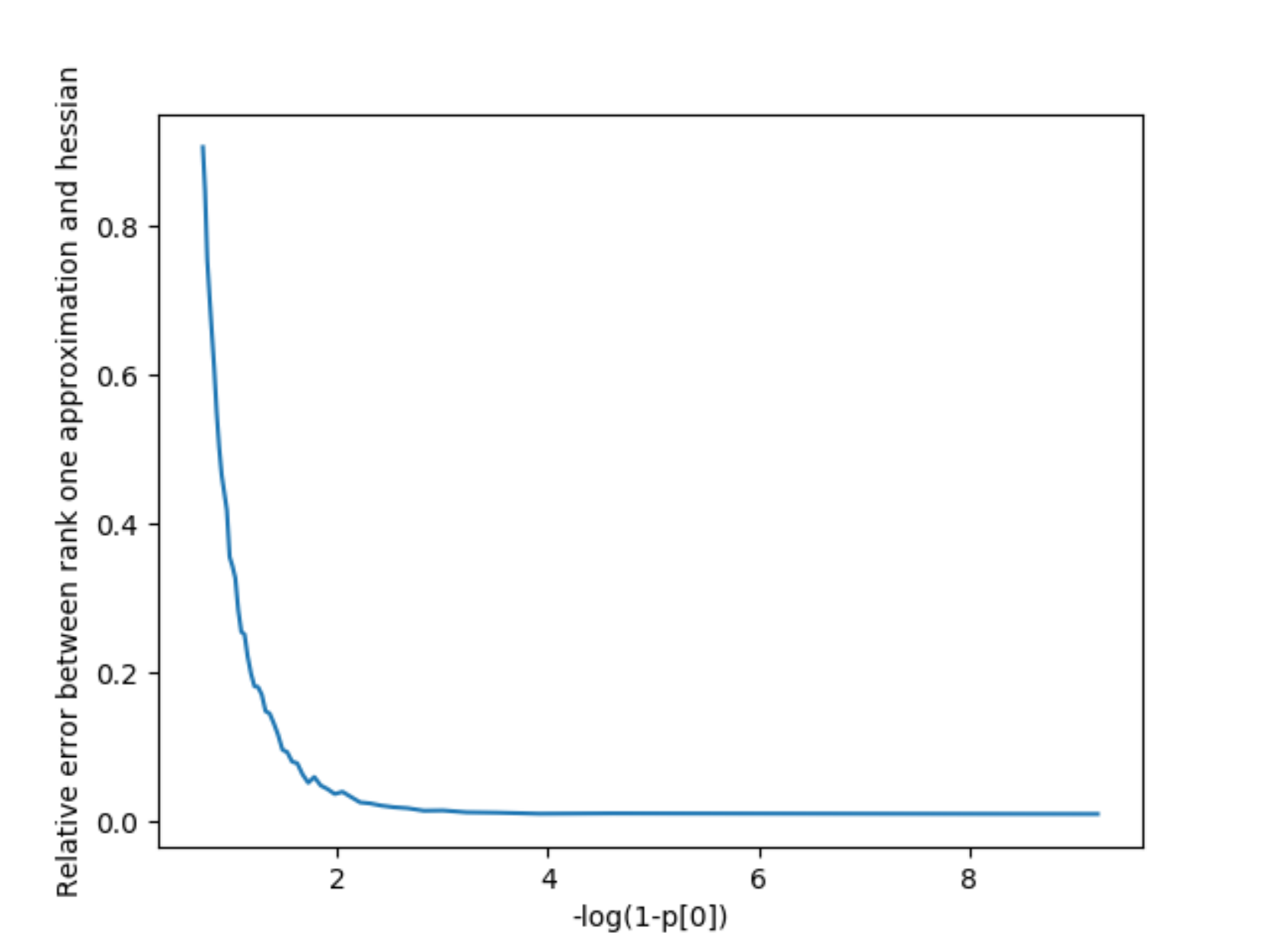}}
		\caption{The relative error between a rank one approximation of the Hessian and the true Hessian as the probability of the predicted class grows. We use a log scale and denote the predicted probability as p[0].}
		\label{fig:rel_errors_vs_eps}
	\end{center}
\end{figure}

Theorems \ref{thm:hessianparallel} and \ref{thm:equivalence} are valid only in the asymptotic regime. A similar analysis in the finite regime is more challenging as it requires the use of perturbation analysis of matrix eigenvalues and eigenvectors. However, we can do a simulation to assess the rate of convergence of the Hessian matrix to such a rank one matrix as the number of classes increases (in Figure \ref{fig:rel_errors_vs_classes}) and the probability of the predicted class tends to 1 (in Figure \ref{fig:rel_errors_vs_eps}). 

\begin{figure}[!ht]
	\begin{center}
		\centerline{\includegraphics[width=\columnwidth]{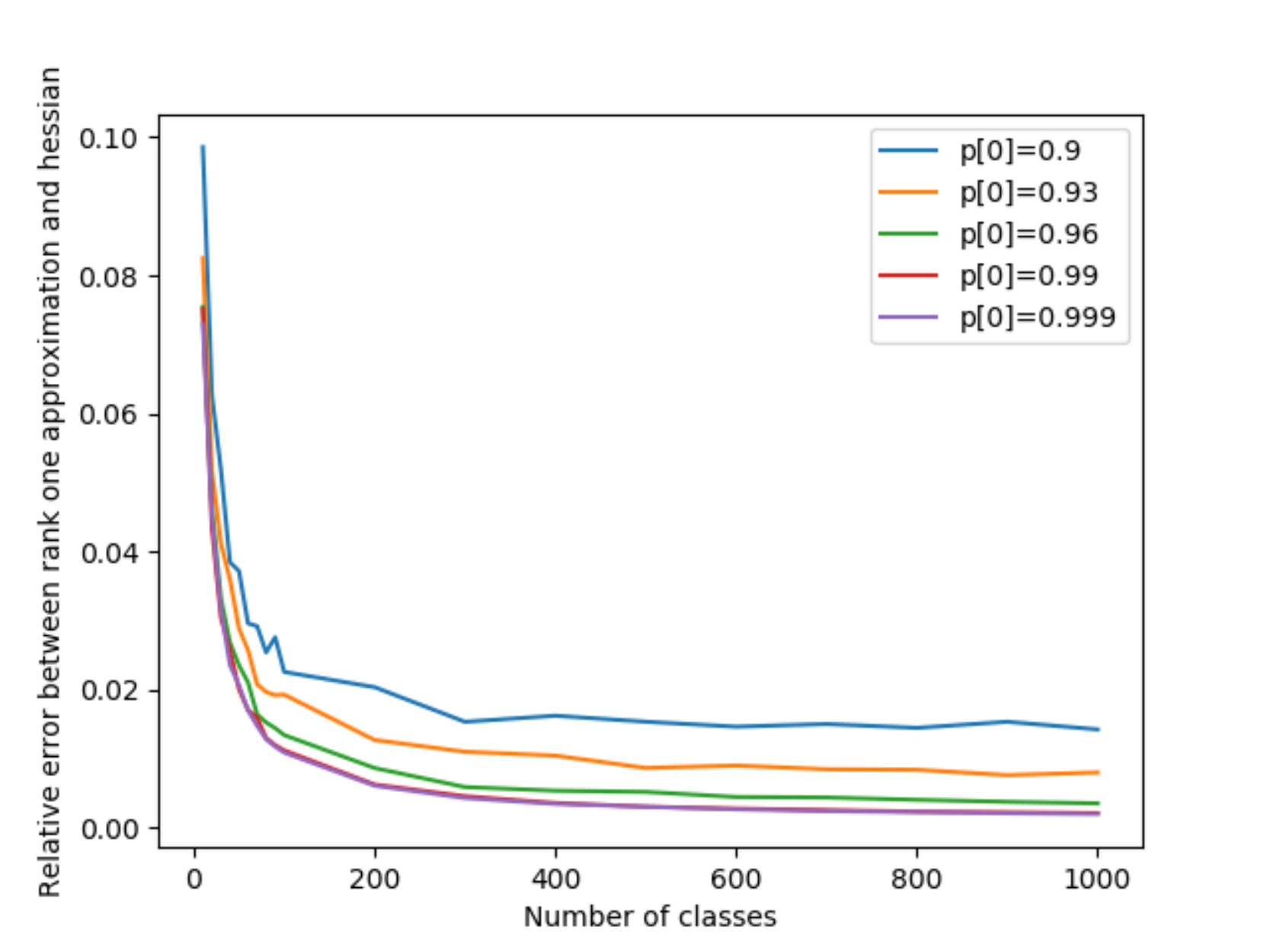}}
		\caption{The relative error between a rank one approximation of the Hessian and the true Hessian when varying the number of classes and the probability of predicted class (denoted by p[0]).}
		\label{fig:rel_errors_vs_classes_vs_eps}
	\end{center}
\end{figure}

For the simulation, we create a linear model $\hby = \bWW^{T}\bx + \bb$ where $\bWW$ and $\bb$ are initialized to random values. Since Theorem \ref{thm:hessianparallel} does not assume a trained network, our analysis is valid even with randomly initialized values of $\bWW$ and $\bb$.
Let the class probabilities be denoted by $\bp$, the number of classes by c and the label vector by $\by$. We again use $\bg_{\bx}$ and $\bH_{\bx}$ as defined in \eqref{eq:gradient} and \eqref{eq:hessian}, respectively.
Without loss of generality, assume that the first class is the one with maximum probability. Thus we create a probability vector $\bp=[1-(c-1)\eps,\ \eps,\ \dotsc,\ \eps]$ and a prediction vector $\by=[1.,\ 0.,\ \dotsc,\ 0.]$. Using Proof \ref{proof:hessianparallel}, we have that the Hessian of this model can be approximated using:
$$\bH_{\bx} \approx \frac{\bg_{\bx}\bg_{\bx}^T}{\eps(c-1)}$$
$$\bg_{\bx} = \bWW(\bp - \by)$$
For Figure \ref{fig:rel_errors_vs_classes}, we fix the probability of predicted class to be 0.9999 (we call it $\bp[0]$) and vary c from 10 to 1000. Thus, note that $\eps$ varies as c varies and is given by $\eps=\frac{1-\bp[0]}{c-1}$. For Figure \ref{fig:rel_errors_vs_eps}, we fix number of classes to be 100 and vary $\eps$ from $5e-3$ to $1e-6$ (an interval length of $1e-6$). Similarly, Figure \ref{fig:rel_errors_vs_classes_vs_eps} shows a comparison between the relative error as the number of classes and probability are both varied. We observe that the relative error converges quickly as a function of both the number of classes and -log(1-$\bp[0]$).

\subsection{Comparing CASO and CAFO for ReLU networks}\label{subsec:sm_low_confidence}
\begin{figure}[!ht]
	\begin{center}
		\centerline{\includegraphics[width=\columnwidth]{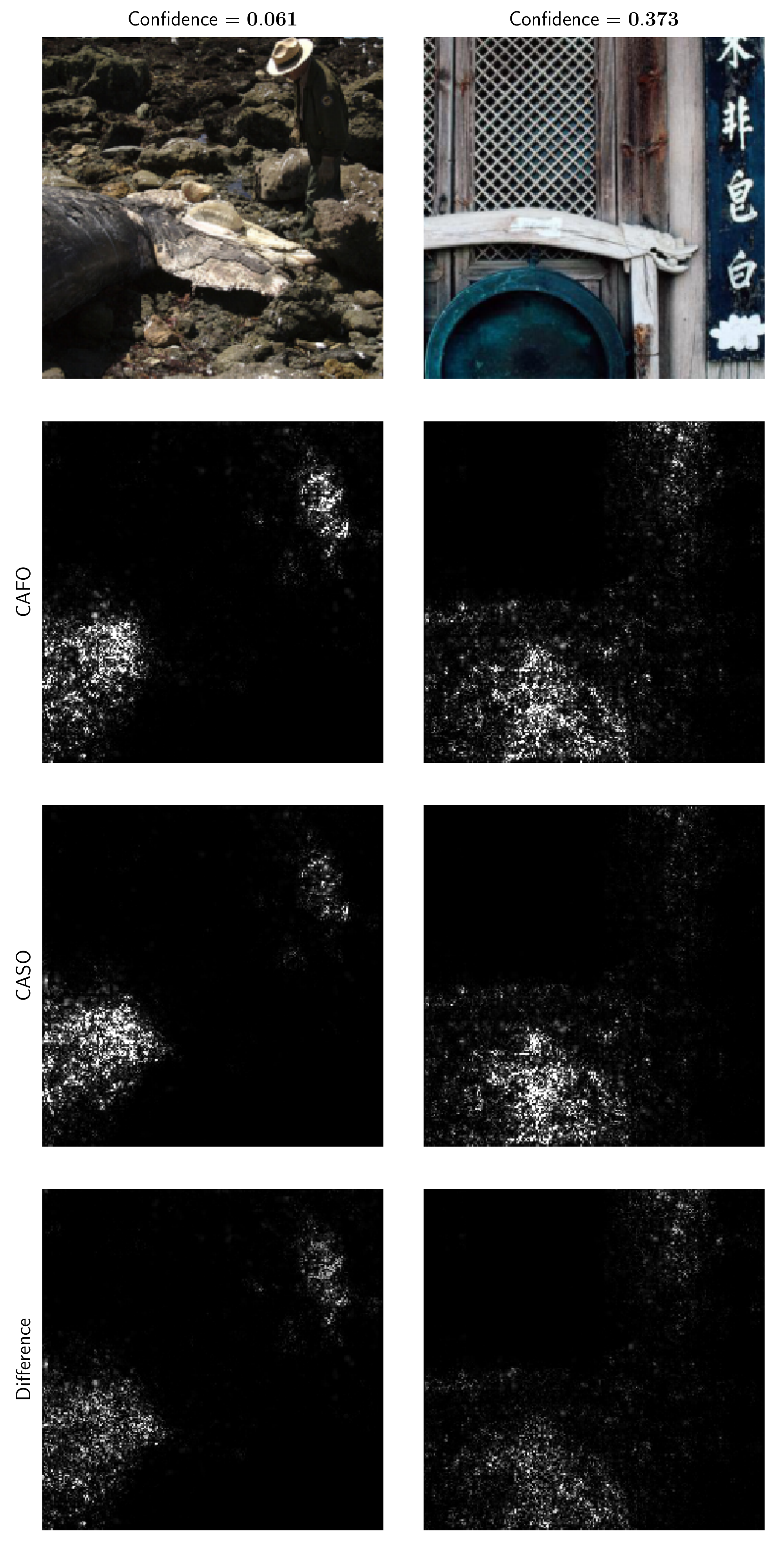}}
		\caption{CASO and CAFO interpretations for low confidence examples for a network with ReLU activations.}
		\label{fig:figure_2_sm_relu}
	\end{center}
\end{figure}

We show additional examples (for relu networks) with low confidence in the predicted class in Figure \ref{fig:figure_2_sm_relu}. The interpretations produced by CASO and CAFO are qualitatively different.\\

\subsection{Experiments with General Non-linearities}\label{subsec:nonrelu_networks}
\begin{figure}[!h]
	\begin{center}
		\centerline{\includegraphics[width=\columnwidth]{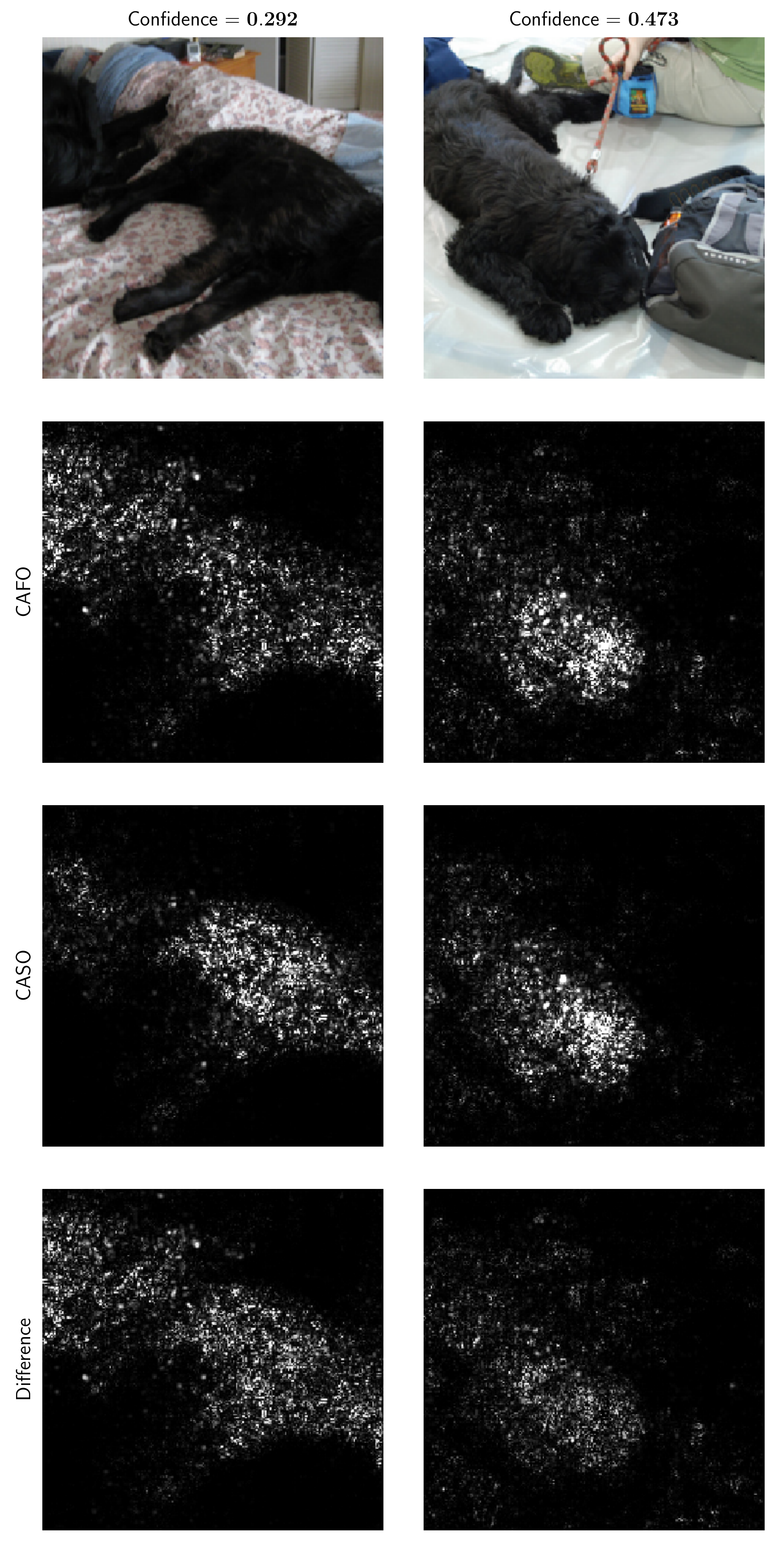}}
		\caption{CASO and CAFO interpretations for low confidence predictions using a model that is not piecewise linear (SE-Resnet-50).}
		\label{fig:figure_2_nonrelu}
	\end{center}
\end{figure}

\begin{figure}[!h]
	\begin{center}
		\centerline{\includegraphics[width=\columnwidth]{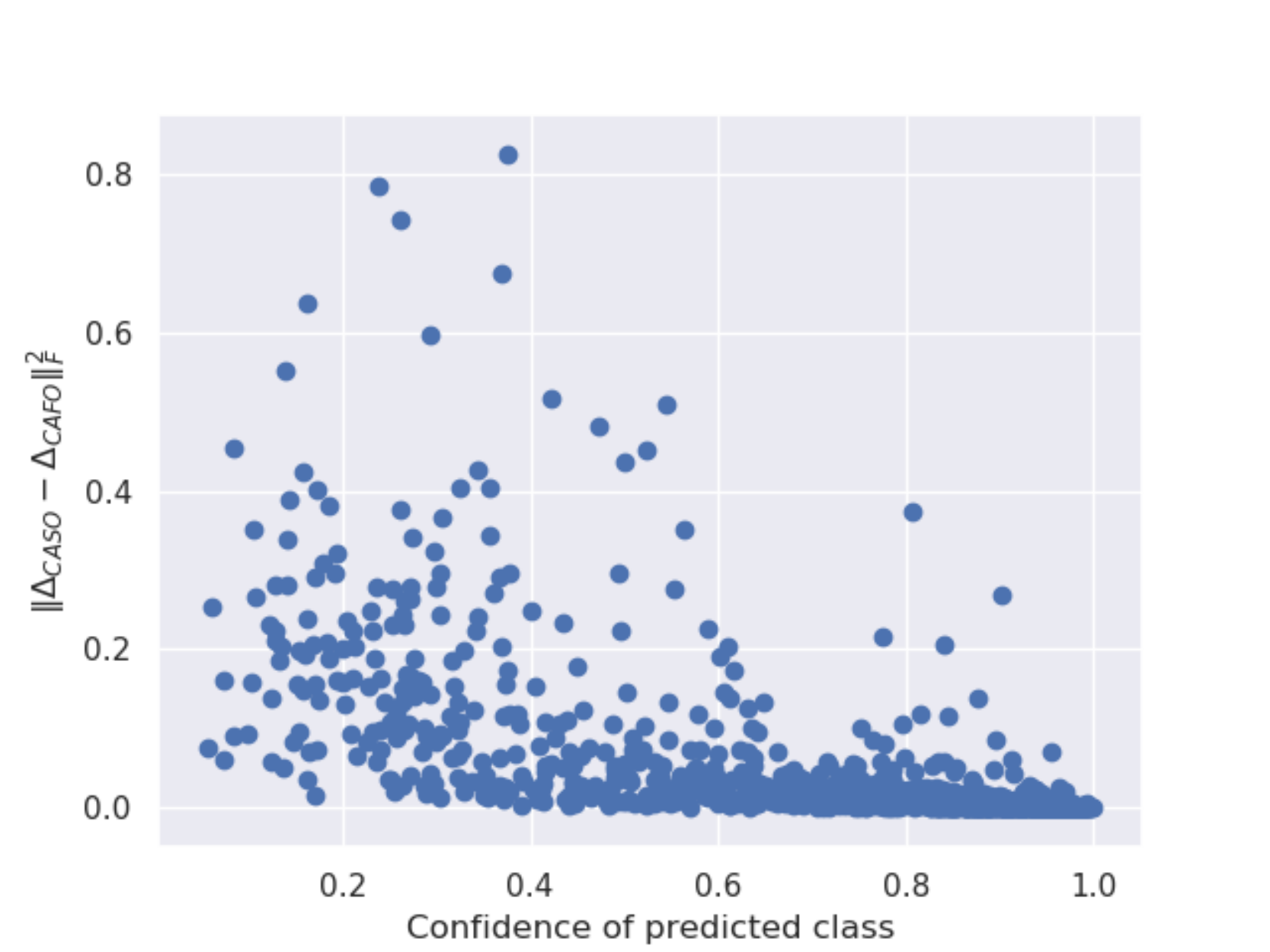}}
		\caption{Scatter plot showing the Frobenius norm difference between CASO and CAFO (after normalizing both vectors to have the same $L_2$ norm) for a network which is not piecewise linear (SE-Resnet-50).}
		\label{fig:conf_frob_nonrelu}
	\end{center}
\end{figure}

We use a SE-Resnet-50 \cite{Hu_2018_CVPR}, a neural network with sigmoid non-linearities. The sigmoid non-linearity causes the model to no longer be piecewise linear. We generate saliency maps using the same 1000 random samples as in Section  \ref{subsec:hessian_impact}. \\
We plot the Frobenius norm of the difference between CASO and CAFO in Figure~\ref{fig:conf_frob_nonrelu}. We normalize the solutions produced by CASO and CAFO to have the same $L_{2}$ norm before taking the difference. Even though the model is no longer piecewise linear, the empirical results are consistent with Theorem \ref{thm:equivalence} (which only holds for piecewise linear networks). \\
To observe the difference between CAFO and CASO interpretations, we compare them for two images with low classification confidence in Figure \ref{fig:figure_2_nonrelu}. The interpretations produced by CASO and CAFO are qualitatively different.

\section{Additional Details on Experiments}
\subsection{Details on Experiments Reported in Figure \ref{fig:scatter-hessian}}
Current autograd software does not support fast eigenvalue decomposition of a matrix in a batched setting. This makes computing the exact hessian inefficient when interpreting numerous samples. For the purposes of this experiment, we use proximal gradient descent to compute the interpretations $\Delta_{CASO}$ and $\Delta_{CAFO}$, even though the parameter $\lambda_{1}$ is set to zero. Details for the other hyperparameters are given in Table \ref{tab:figure_1}.
\begin{table}[!h]
	\caption{Hyper-parameter details for Figure \ref{fig:scatter-hessian}}
	\centering
	\begin{tabular}{c c}
		\hline
		Parameter & Configuration \\   
		\hline
		\hline
		$\lambda_{1}$ & $0$ \\
		$\lambda_{2}$ threshold & $20$ \\
		Optimizer & Proximal Gradient Descent \\
		Network architecture & Resnet-50 \\
		Batch size & $32$\\
		Power method iterations & $10$\\
		Gradient descent iterations & $10$\\
		Backtracking decay factor & 0.5\\
		Initialization & Zero\\
		\hline
	\end{tabular}
	\label{tab:figure_1}
\end{table}

\subsection{Details on Experiments Reported in Figure \ref{fig:lambda1_sweep}}
Details of the hyperparameters used in Figure~\ref{fig:lambda1_sweep} are shown in Table \ref{tab:figure_3}.
\begin{table}[!h]
	\caption{Hyper-parameter details for Figure \ref{fig:lambda1_sweep}}
	\centering
	\begin{tabular}{c c}
		\hline
		Parameter & Configuration \\   
		\hline
		\hline
		\multirow{3}{*}{$\lambda_{1}$ values}&0, 10$^{-5}$, 10$^{-4}$, 10$^{-3}$,\\
		&6.25$\times$10$^{-3}$, 1.25$\times$10$^{-2}$, \\
		&2.5$\times$10$^{-2}$, 5$\times$10$^{-2}$\\
		$\lambda_{2}$ threshold & $20$ \\
		Optimizer & Proximal Gradient Descent \\
		Network architecture & Resnet-50 \\
		Batch size & $32$\\
		Power method iterations & $10$\\
		Gradient descent iterations & $10$\\
		Backtracking decay factor & 0.5\\
		Initialization & Zero\\
		\hline
	\end{tabular}
	\label{tab:figure_3}
\end{table}

\subsection{Details on Experiments Reported in Figure \ref{fig:comp_existing}}
Details of the hyperparameters used in Figure~\ref{fig:comp_existing} are shown in Table \ref{tab:figure_4}.
\begin{table}[!h]
	\caption{Hyper-parameter details for Figure \ref{fig:comp_existing}}
	\centering
	\begin{tabular}{c c}
		\hline
		Parameter & Configuration \\   
		\hline
		\hline
		\multirow{3}{*}{$\lambda_{1}$ values}&0, 10$^{-5}$, 10$^{-4}$, 10$^{-3}$,\\
		&6.25$\times$10$^{-3}$, 1.25$\times$10$^{-2}$, \\
		&2.5$\times$10$^{-2}$, 5$\times$10$^{-2}$\\
		$\lambda_{2}$ threshold & $20$ \\
		Optimizer & Proximal Gradient Descent \\
		Network architecture & Resnet-50 \\
		Batch size & $32$\\
		Power method iterations & $10$\\
		Gradient descent iterations & $10$\\
		Backtracking decay factor & 0.5\\
		Number of samples & 32\\
		Stddev of Random samples & 0.15\\
		Initialization & Zero\\
		\hline
	\end{tabular}
	\label{tab:figure_4}
\end{table}

\section{Comparison with existing methods}\label{sec:SM_existing_comp}
\begin{figure*}[!h]
	\begin{center}
		\centerline{\includegraphics[width=\textwidth, trim={0cm 0cm 0cm 0cm}, clip]{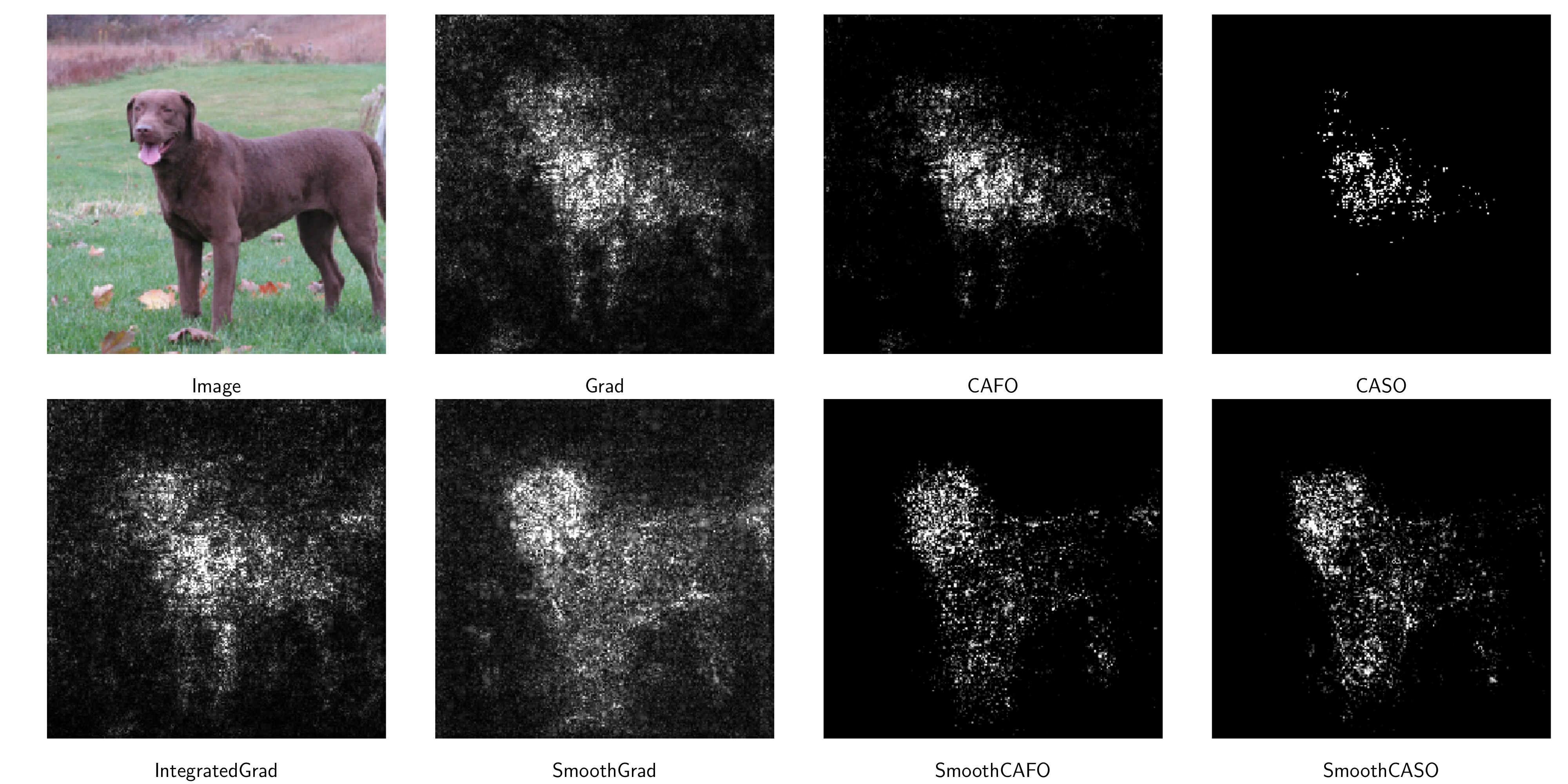}}
	\end{center}
	\caption{}
	\label{fig:comp_existing_1}
\end{figure*}

\begin{figure*}[!h]
	\begin{center}
		\centerline{\includegraphics[width=\textwidth, trim={0cm 0cm 0cm 0cm}, clip]{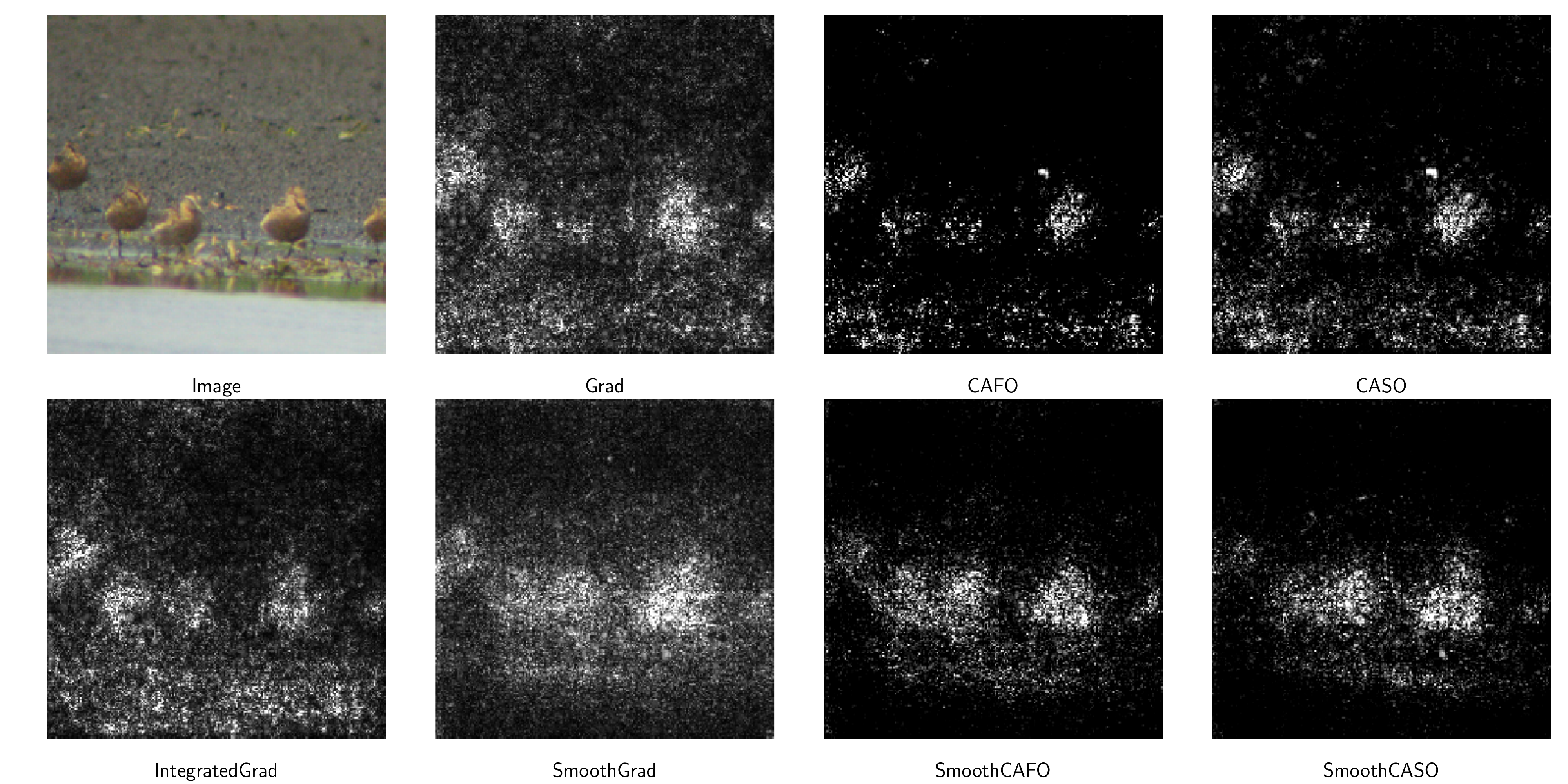}}
	\end{center}
	\caption{}
	\label{fig:comp_existing_2}
\end{figure*}

\begin{figure*}[!h]
	\begin{center}
		\centerline{\includegraphics[width=\textwidth, trim={0cm 0cm 0cm 0cm}, clip]{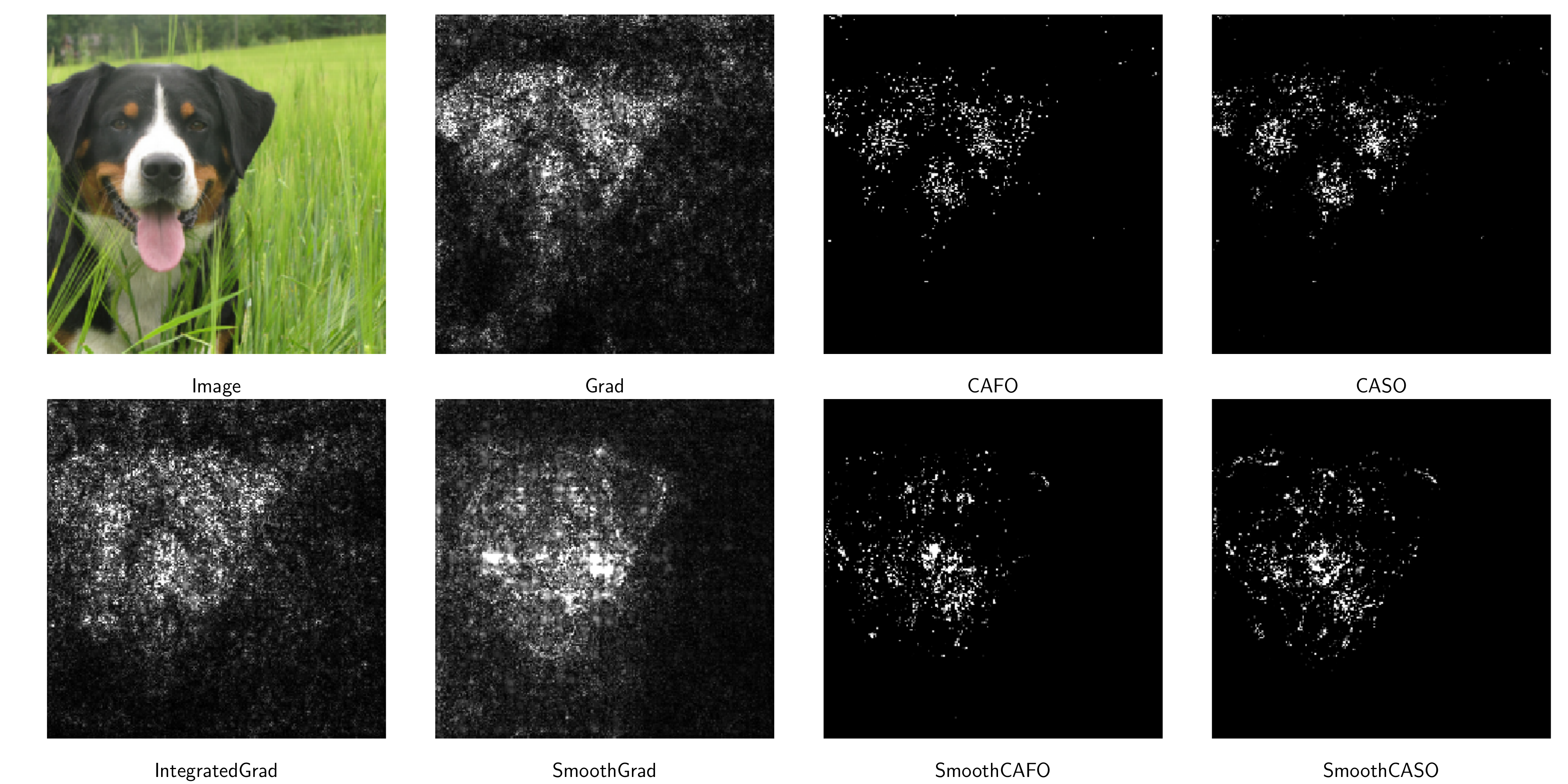}}
	\end{center}
	\caption{}
	\label{fig:comp_existing_3}
\end{figure*}

\begin{figure*}[!h]
	\begin{center}
		\centerline{\includegraphics[width=\textwidth, trim={0cm 0cm 0cm 0cm}, clip]{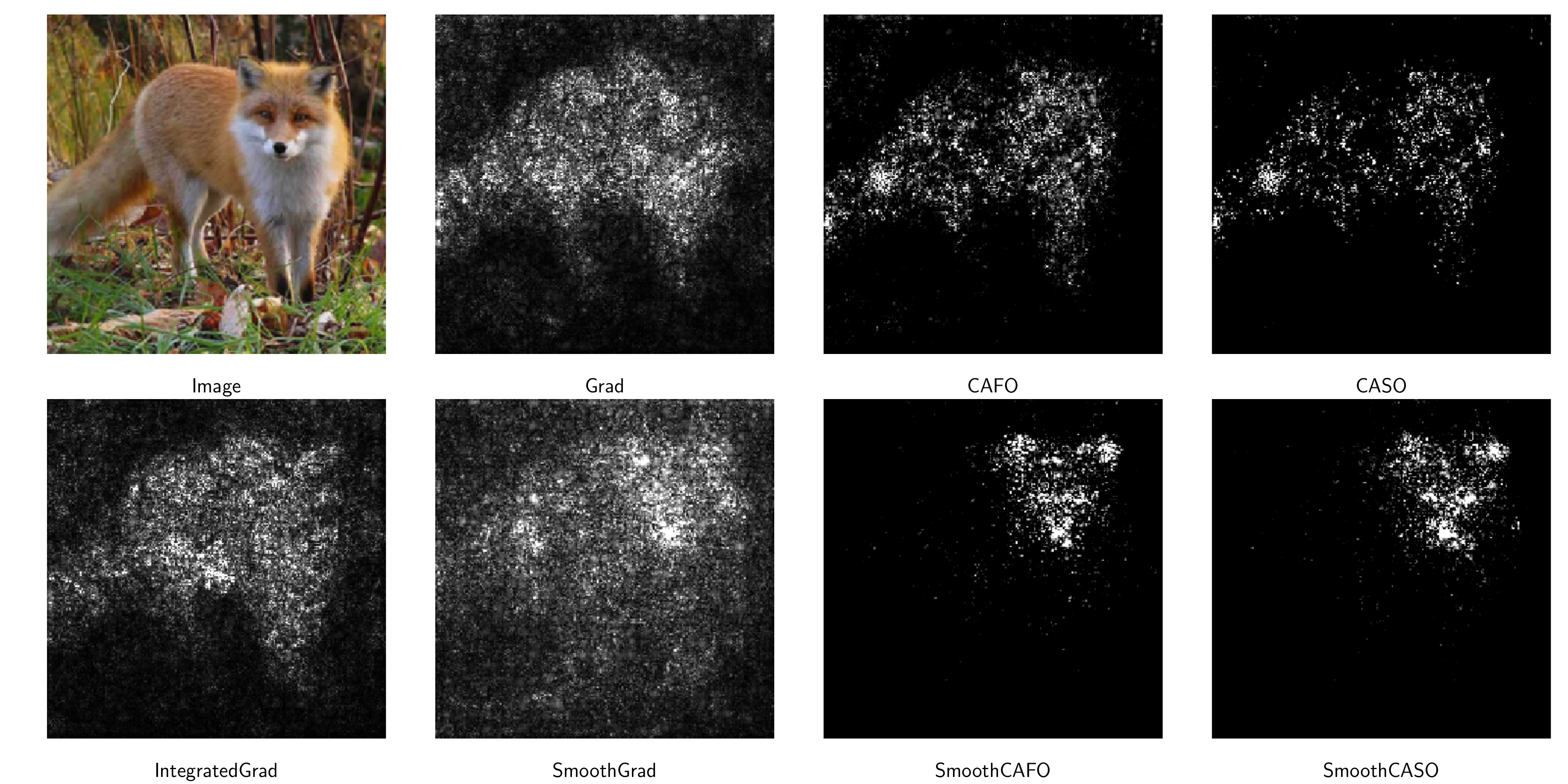}}
	\end{center}
	\caption{}
	\label{fig:comp_existing_4}
\end{figure*}

\begin{figure*}[!h]
	\begin{center}
		\centerline{\includegraphics[width=\textwidth, trim={0cm 0cm 0cm 0cm}, clip]{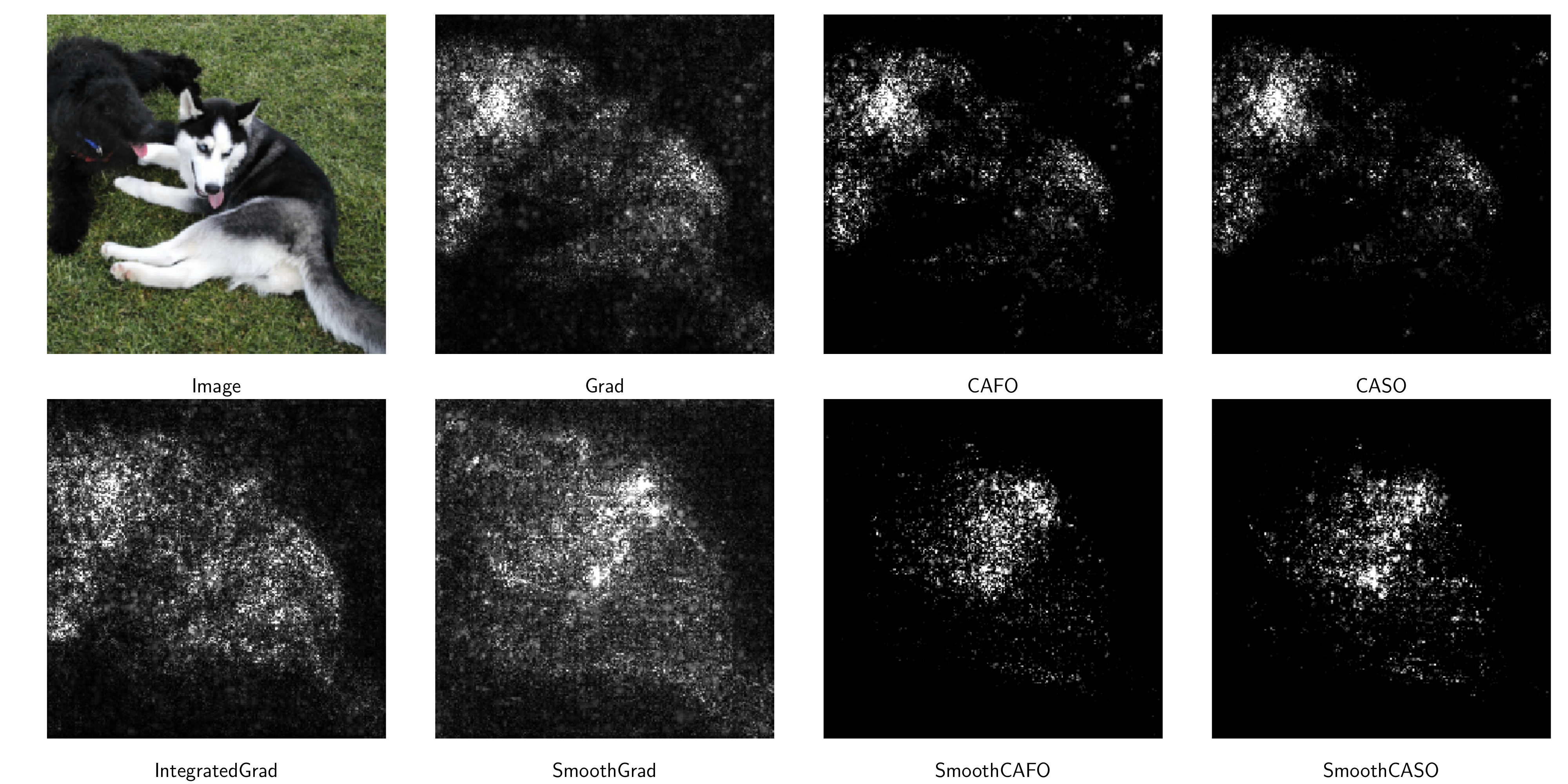}}
	\end{center}
	\caption{}
	\label{fig:comp_existing_5}
\end{figure*}

\begin{figure*}[!h]
	\begin{center}
		\centerline{\includegraphics[width=\textwidth, trim={0cm 0cm 0cm 0cm}, clip]{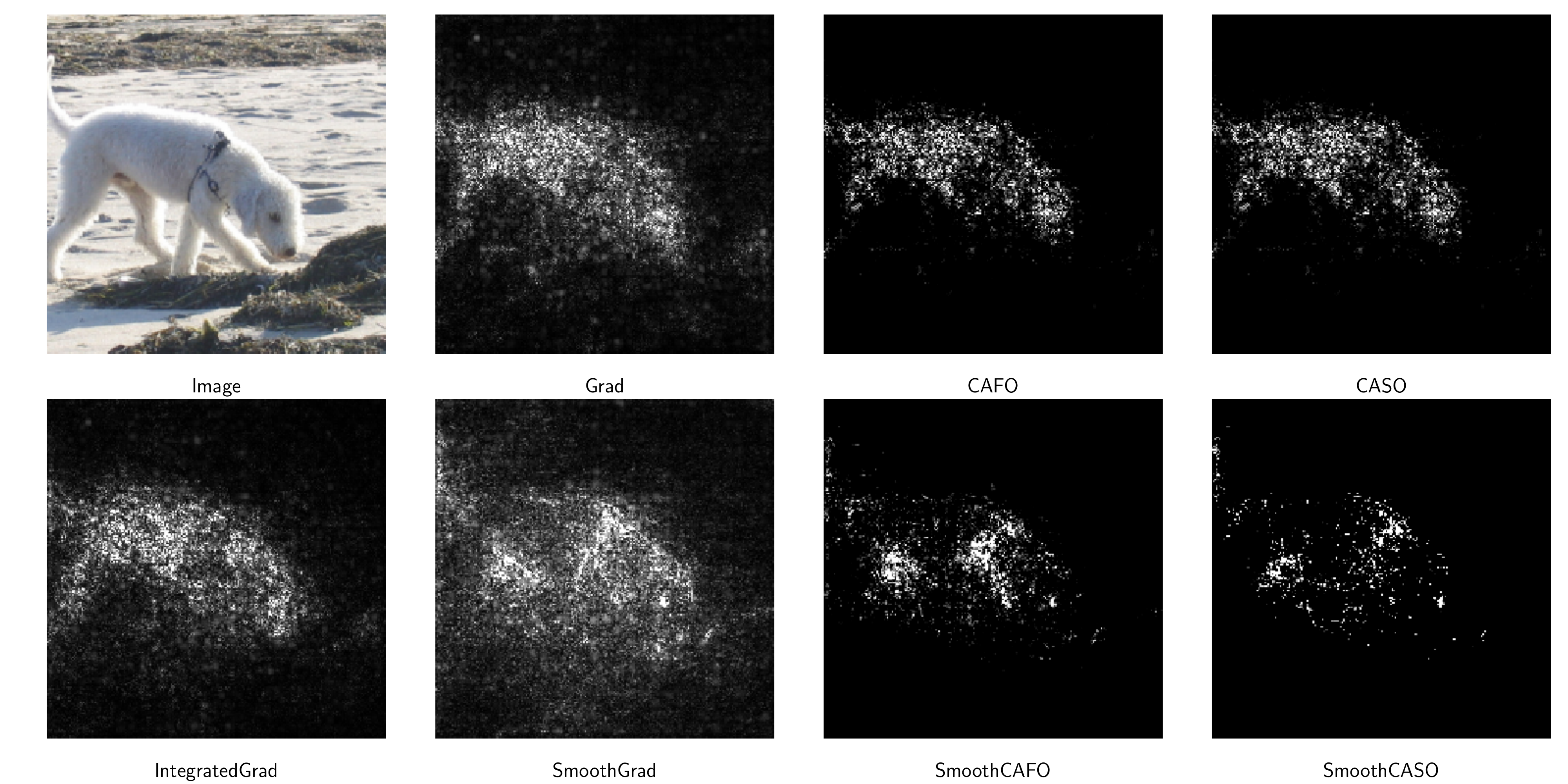}}
	\end{center}
	\caption{}
	\label{fig:comp_existing_6}
\end{figure*}

Figures~\ref{fig:comp_existing_1}--\ref{fig:comp_existing_6} provide further examples of our interpretation method with existing techniques.

\end{document}